\documentclass[11pt,a4paper]{article}

\usepackage[T1]{fontenc}
\usepackage[utf8]{inputenc}
\usepackage{CJKutf8}
\usepackage{times}

\usepackage[margin=2.4cm,headheight=38pt,headsep=18pt,footskip=28pt]{geometry}
\usepackage{setspace}
\usepackage[table,dvipsnames]{xcolor}
\usepackage{colortbl}
\usepackage{graphicx}
\usepackage{booktabs}
\usepackage{pifont}

\newcommand{\cmark}{\ding{51}}
\usepackage{pifont}
\usepackage{array}
\usepackage{multirow}
\usepackage{tabularx}
\usepackage{wrapfig} 
\usepackage{subcaption}
\usepackage{booktabs}   
\usepackage{adjustbox}     
\usepackage{placeins}      
\usepackage{amssymb}       

\usepackage{amsmath,amssymb}

\definecolor{AMapBlue}{HTML}{48B8F0}
\definecolor{AMapDeepBlue}{HTML}{0877C9}
\definecolor{AMapSky}{HTML}{EAF8FF}
\definecolor{AMapLine}{HTML}{9EDBFA}
\definecolor{TextBlack}{HTML}{20242A}

\usepackage[numbers,sort&compress]{natbib}

\usepackage{enumitem}
\setlist{nosep,leftmargin=1.5em}
\usepackage{caption}
\usepackage{fancyhdr}
\usepackage{lastpage}
\usepackage{titlesec}
\usepackage[most]{tcolorbox}
\usepackage{mdframed}
\usepackage[colorlinks=true,linkcolor=AMapDeepBlue,citecolor=AMapDeepBlue,urlcolor=AMapDeepBlue]{hyperref}

\newcommand{\reportname}{TriFold}
\newcommand{\reportlogo}{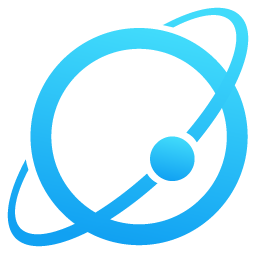}
\newcommand{\reportlogoheight}{18pt}

\newcommand{\orgname}{AMAP CV Lab \textbar{} Alibaba Group}
\newcommand{\reportcenter}[1]{\raisebox{\dimexpr(\depth-\height)/2\relax}{#1}}
\newlength{\reporttextbaselineoffset}
\newcommand{\reporttextalign}[1]{\raisebox{\reporttextbaselineoffset}{#1}}

\renewcommand{\headrulewidth}{1.2pt}
\renewcommand{\footrulewidth}{0.8pt}
\renewcommand{\headrule}{\hbox to\headwidth{\color{AMapBlue}\leaders\hrule height \headrulewidth\hfill}}
\renewcommand{\footrule}{\hbox to\headwidth{\color{AMapLine}\leaders\hrule height \footrulewidth\hfill}}

\titleformat{\section}
  {\Large\bfseries\color{AMapDeepBlue}}
  {\thesection}{0.75em}{}
  [\vspace{0.32em}{\color{AMapLine}\titlerule[0.6pt]}]
\titleformat{\subsection}
  {\large\bfseries\color{AMapDeepBlue}}
  {\thesubsection}{0.65em}{}
\titleformat{\subsubsection}
  {\normalsize\bfseries\color{AMapDeepBlue}}
  {\thesubsubsection}{0.55em}{}
\titlespacing*{\section}{0pt}{1.55em plus 0.25em minus 0.1em}{0.95em}
\titlespacing*{\subsection}{0pt}{1.25em plus 0.2em minus 0.1em}{0.55em}
\titlespacing*{\subsubsection}{0pt}{1.05em plus 0.15em minus 0.08em}{0.45em}

\newtcolorbox{abstractbox}{
  enhanced,
  breakable,
  colback=AMapSky,
  colframe=AMapBlue,
  coltitle=white,
  colbacktitle=AMapBlue,
  title=Abstract,
  fonttitle=\Large\bfseries,
  halign title=center,
  toptitle=5pt,
  bottomtitle=5pt,
  fontupper=\large,
  boxrule=0.8pt,
  arc=1.5mm,
  left=12pt,
  right=12pt,
  top=10pt,
  bottom=10pt,
  before skip=1.2em,
  after skip=1.4em
}

\newcommand{\modelname}{TriFold}
\newmdenv[
  backgroundcolor=gray!8,
  linewidth=0pt,
  innerleftmargin=10pt,
  innerrightmargin=10pt,
  innertopmargin=5pt,
  innerbottommargin=5pt,
  skipabove=5pt,
  skipbelow=5pt
]{cnblock}

\definecolor{DimAction}{HTML}{0877C9}
\definecolor{DimVisual}{HTML}{2E9E5B}
\definecolor{DimPhysics}{HTML}{E07B00}
\definecolor{DimMemory}{HTML}{8E44AD}

\usepackage{needspace}
\let\reportsection\section
\renewcommand{\section}{\Needspace{7\baselineskip}\reportsection}
\begin{document}
\thispagestyle{fancy}
\begin{center}
{\bfseries\color{AMapDeepBlue}\LARGE TriFold: Folding irregular 3D Gaussians into structured latents for 3D Scene Generation\par}
\vspace{0.2em}
{\large\color{black}
Yun Wang\textsuperscript{\ensuremath{2,*,\S}}\quad Chiyu Wang\textsuperscript{\ensuremath{1,*,\dagger}}\quad Fan Jiang\textsuperscript{\ensuremath{1,*,\ddagger}}\quad Ziyu Zhu\textsuperscript{\ensuremath{1}}\\
Zhaoxu Sun\textsuperscript{\ensuremath{1}}\quad Guangchi Fang\textsuperscript{3}\quad Mu Xu\textsuperscript{1}\quad Dapeng Wu\textsuperscript{\ensuremath{2}}\par}
\vspace{-0.1em}
{\small\color{black}\textsuperscript{1}AMAP, Alibaba Group\qquad\textsuperscript{2}City University of Hong Kong \qquad\textsuperscript{3}Hong Kong Polytechnic University}
\end{center}
\vspace{-1em}
\begingroup
\renewcommand{\thefootnote}{}
\makeatletter
\long\def\@makefntext#1{\noindent#1}
\makeatother
\footnotetext[0]{\textsuperscript{*}Equal contribution.\quad
\textsuperscript{\ensuremath{\dagger}}Project leader.\quad
\textsuperscript{\ensuremath{\ddagger}}Corresponding author.\\
\textsuperscript{\S}Work done during internship at AMAP, Alibaba Group.}
\endgroup
\begin{abstractbox}
Recent progress in 3D generation has motivated growing interest in compact generative representations of complex 3D scenes. For native 3D Gaussian scenes, however, learning such representations remains challenging due to widely varying primitive counts, nonuniform spatial density, and differences in geometry and appearance even among spatially neighboring primitives.
In this work, we introduce \textbf{\modelname{}}, a native 3D Gaussian VAE that encodes Gaussian scenes into a fixed-layout triplane latent and jointly reconstructs their geometry and appearance.
The key insight is to impose structural regularity on the latent space while preserving the spatial adaptivity of the underlying Gaussians.
During encoding, we use cell-wise learned queries to aggregate projected Gaussians into multiple slots per cell, reducing the information loss caused by single-feature pooling.
During decoding, occupancy prediction identifies active regions, where query-based set prediction and candidate selection reconstruct local Gaussian sets with spatially varying primitive counts.
Building on this representation, we train a latent flow-matching model for conditional scene-level 3DGS generation, enabling point-cloud-conditioned synthesis and single-image-conditioned amodal generation.
Experiments demonstrate that \modelname{} achieves high-fidelity reconstruction and generation from a single image that provides only a partial observation of the scene, validating its effectiveness for native
3D Gaussian representation learning.
\href{https://fantasy-amap.github.io/TriFold/#}{https://fantasy-amap.github.io/TriFold}

\end{abstractbox}

\section{Introduction}
Amodal 3D scene generation aims to synthesize complete environments from partial observations, enabling exploration and embodied interaction beyond captured views~\citep{devries2021unconstrained,bautista2022gaudi,meng2025lt3sd,chen2026nova3r,ngo2026volfill,sargent2024zeronvs,jin2024fires,weber2024nerfiller,dai2020sgnn}. This requires a representation that captures complex 3D structure while remaining tractable for generative modeling.
Recent scene-level systems use compositional neural fields, progressive expansion, sparse volumes, or latent triplanes to model room-scale and unbounded environments~\citep{chen2023scenedreamer,hollein2023text2room,ren2024xcube,wu2024blockfusion,ju2024diffindscene,zhang2024berfscene,zhang20243d}.
Recent advances increasingly rely on structured latent representations that reorganize irregular 3D signals into formats compatible with generative models~\citep{wang2023rodin,li2025triposg,lai2025hunyuan3d25highfidelity3d,xiang2025trellis,xiang2026native}. 
However, extending this paradigm to native scene-level 3D Gaussian Splatting (3DGS)~\citep{kerbl2023gaussians} poses distinct representation challenges. 
Scene-level Gaussian distributions are heterogeneous: globally, scenes can differ by orders of magnitude in primitive count and exhibit highly nonuniform spatial density; locally, spatially adjacent Gaussians suffer from severe attribute heterogeneity: neighboring primitives often differ sharply in geometry and appearance~\citep{lee2026optimized}.
These variations motivate a latent representation that is structured for generative modeling while preserving scene-specific Gaussian cardinality, spatial distributions, and fine-grained local attributes.

\begin{figure}[t]
    \centering
    \includegraphics[width=1\linewidth]{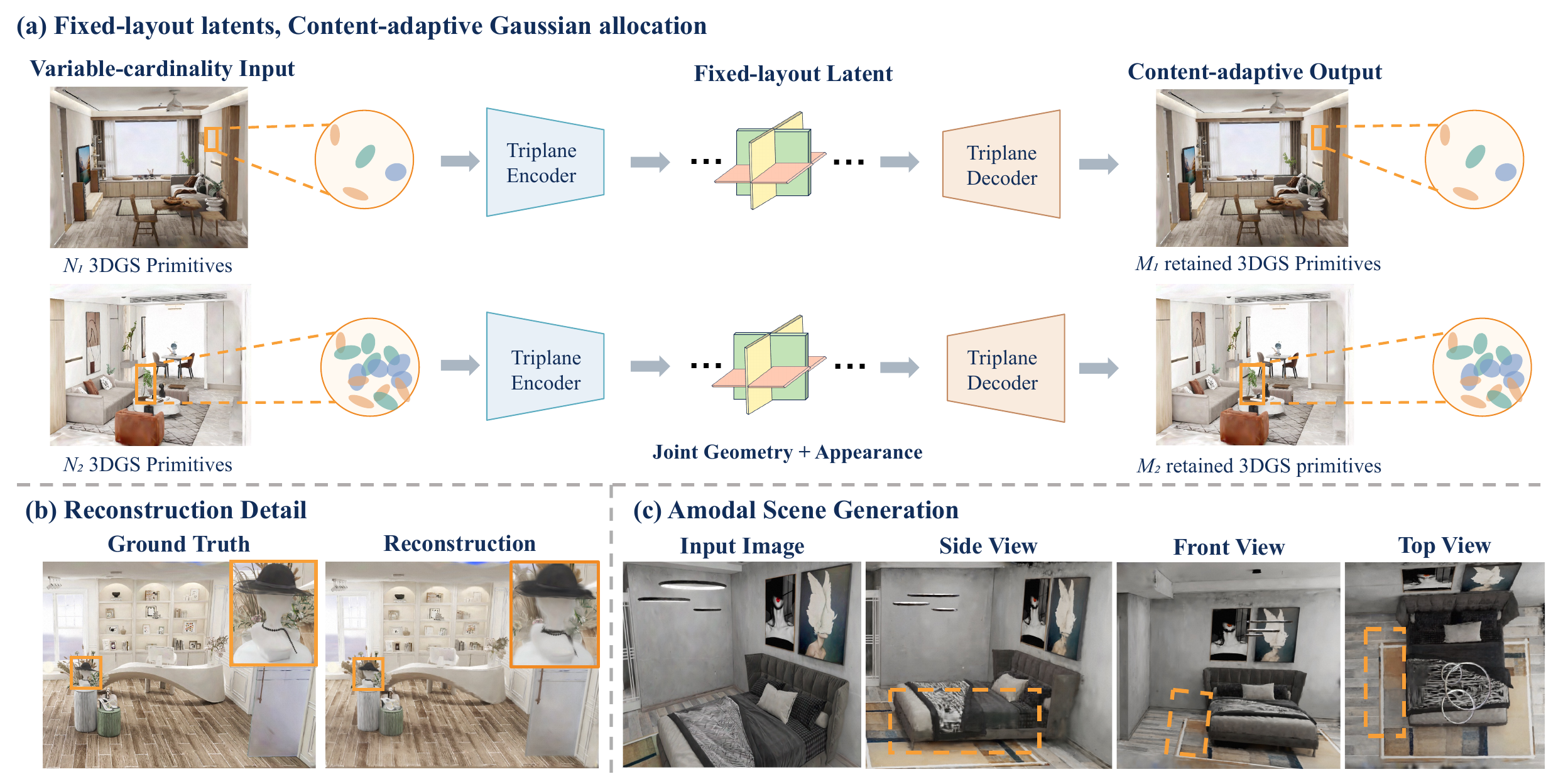}
   \vspace{-0.6cm}
    \caption{
    \textbf{Overview of \modelname{}.}
    (a) \modelname{} encodes variable-cardinality Gaussian scenes into fixed-layout triplane latents that jointly represent geometry and appearance, and decodes content-adaptive Gaussian sets with variable local densities.
    (b) The learned representation supports detailed scene reconstruction.
    (c) Our single-image-conditioned amodal generation preserves observed content while inferring plausible geometry and appearance for unobserved regions.
    }
    \vspace{-0.4cm}
    \label{fig:teaser}
\end{figure}

Existing methods reconcile native Gaussian scenes with latent generative modeling through diverse structural paradigms.
Quantized-grid methods, such as L3DG~\citep{roessle2024l3dg} and GaussianGPT~\citep{vonluetzow2026gaussiangpt}, discretize Gaussians into voxel-aligned representations with predefined local capacity before vector quantization. 
Such fixed per-voxel budgets can underrepresent dense regions while wasting capacity in sparse ones.
Canonical-token methods instead regularize globally. Can3Tok~\citep{gao2025can3tok} resamples each scene into a fixed-cardinality primitive set, inadvertently imposing a uniform representation budget irrespective of structural complexity.
Sparse latent methods offer topological flexibility by anchoring features to dynamic active voxels. 
TRELLIS~\citep{xiang2025trellis}, for example, constructs sparse latents and decodes geometry via a cascaded pipeline. 
While highly effective for bounded, object-centric assets with comprehensive visual coverage, extending this paradigm to expansive scene-level 3DGS is non-trivial: complex occlusions across broad scenes complicate exhaustive multiview feature anchoring, while scaling sparse voxel grids and multi-stage pipelines over unconstrained volumes introduces notable computational and representation overhead.
Finally, continuous-field methods, such as DiffGS~\citep{zhou2024diffgs}, reparameterize Gaussians as continuous spatial fields via feature interpolation; yet, this bias toward spatial smoothness can conflict with the high-frequency variations in attributes such as scale and rotation across neighboring primitives.
Consequently, an open challenge remains: how can we provide generative backbones with a regular, fixed-layout latent space, while preserving the unconstrained, variable-length nature of native Gaussian representations at both the input and output?

To bridge this gap, we present \textbf{\modelname{}}, a variational autoencoder that operates directly on native 3D Gaussian primitives for scene-level generative modeling. Our key design principle is to impose structural regularity in the latent space that jointly represents geometry and appearance, and preserve the flexible, content-adaptive organization of native Gaussian primitives. Specifically, the encoder maps variable-cardinality, non-uniformly distributed Gaussian sets into fixed-layout triplane latents. It first extracts context-aware primitive features with a point transformer. Since multiple features may project to the same triplane cell, we use multiple learned queries per cell to aggregate local information. 
Then, we use Cross-plane attention to integrate information across planes through shared Gaussian projections, followed by a variational bottleneck that produces triplane latents.

To recover native 3DGS from this structured latent, our decoder adopts a set-based decoding paradigm without rigid 3D discretization. An occupancy predictor first localizes active regions, while an optimal-transport-guided set decoder dynamically predicts and filters Gaussian candidates within each region. 
Because the predefined query count serves only as a local capacity upper bound rather than a fixed allocation, the spatial allocation and primitive cardinality remain adaptive to scene content. We supervise Gaussian centers through optimal-transport matching, while the remaining geometric and appearance attributes are constrained by a full-scene rendering loss. Together, these objectives enable accurate Gaussian reconstruction with high rendering fidelity, providing a regular tensor interface for generative modeling while retaining the flexible organization of native 3DGS.

Building on this structured latent space, we train a latent flow-matching model for conditional scene-level 3DGS generation. We consider two conditioning settings: point-cloud-conditioned synthesis from complete spatial observations and single-image-conditioned amodal generation. In the latter, a single image provides only a partial observation of the scene, requiring the learned scene priors to preserve observed content while inferring plausible geometry and appearance in unobserved regions. Experiments demonstrate high-fidelity reconstruction and generation in both settings, demonstrating the effectiveness of structured triplane latents for native 3D Gaussian scene generation.

\begin{itemize}

\item \textbf{A native scene-level 3D Gaussian variational autoencoder.}
We introduce \modelname{}, which maps variable-cardinality Gaussian scenes into a fixed-layout triplane latent and decodes them back into content-adaptive native 3DGS. 

\item \textbf{Unified geometry-appearance modeling.} Our unified representation jointly encodes scene geometry and appearance, and we reconstruct both from the same representation through an occupancy predictor and a query-based set decoder.

\item \textbf{Conditional scene-level 3DGS generation based on \modelname{}.}
We train a latent flow-matching model on the learned triplane latents, enabling both point-cloud-conditioned synthesis and single-image-conditioned amodal generation with plausible geometry and appearance in unobserved regions.

\end{itemize}

\section{Related Work}
\label{sec:related_work}

\paragraph{3D Representations and Generative Models.}
The choice of scene representation shapes the fidelity and scalability of 3D generative models.  Meshes represent explicit surfaces through vertices and faces, while voxel grids discretize space into volumetric cells storing occupancy or other attributes, with dense grids incurring cubic memory growth with resolution~\citep{qi2017pointnet,lorensen1987marchingcubes,kazhdan2006poisson,hou2022iterative,riegler2017octnet}. Neural radiance fields represent scenes as continuous functions and support high-quality novel-view synthesis, but require repeated neural evaluations during rendering~\citep{mildenhall2020nerf,wang2021neus,guo2023streetsurf}. 3DGS instead uses adaptive explicit primitives for differentiable real-time rendering~\citep{kerbl2023gaussians}. 
Representative generative approaches include $\pi$-GAN for implicit radiance
fields~\citep{chan2021pi}, SDFusion for volumetric SDF grids
~\citep{cheng2023sdfusion}, GET3D and MeshGPT for explicit meshes
~\citep{gao2022get3d,siddiqui2024meshgpt}, and 3DShape2VecSet, Michelangelo,
and TripoSG for generation in learned 3D latent spaces
~\citep{zhang20233dshape2vecset,zhao2023michelangelo,li2025triposg}.

These properties make 3DGS attractive for generative modeling. Native 3DGS,
however, represents each scene as an unordered set whose cardinality and
spatial density adapt to scene content. This organization is mismatched with
latent diffusion and flow-matching backbones, which typically operate on
fixed-shape tensors~\citep{rombach2022high,esser2024scaling}. Triplanes offer
a regular layout with a practical balance between spatial expressiveness and
computational efficiency and have been widely used for 3D-aware generation
~\citep{chan2022eg3d,wu2024direct3d}. The central challenge is therefore to map adaptive
Gaussian sets into such fixed-layout latents while retaining their
scene-dependent cardinality and spatial density during reconstruction.

\paragraph{Latent Representations for 3D Gaussians.}
We focus on methods that regularize Gaussian representations or construct
generative 3D latents with Gaussian decoders. This differs from image-conditioned 3DGS methods, which directly predict Gaussian representations from one or more images for reconstruction or
content creation
~\citep{charatan2024pixelsplat,zou2024triplane,tang2024lgm,xu2024grm,poole2023dreamfusion,lin2023magic3d,melaskyriazi2023realfusion,liu2023zero123}.
Within this generative latent setting, existing approaches differ primarily
in how they impose structural regularity.
Grid-based methods impose explicit spatial organization: GaussianCube rearranges a fixed number of Gaussians onto voxel grids through optimal transport, whereas L3DG and GaussianGPT adopt voxel-aligned Gaussian representations with vector-quantized autoencoders~\citep{zhang2024gaussiancube,roessle2024l3dg,vonluetzow2026gaussiangpt}. Canonical-token methods such as Can3Tok convert a scene into a fixed-cardinality token set~\citep{gao2025can3tok}. Sparse approaches retain scene-dependent spatial support: TRELLIS constructs sparse latent features from multiview observations and generates spatial structure and latent features in two stages, whereas GS-Voxel reorganizes fitted Gaussian scenes into sparse voxel structures with scene-dependent support~\citep{xiang2025trellis,qian2026gsvoxel}. Continuous-field methods such as DiffGS decode Gaussian attributes from interpolated spatial features~\citep{zhou2024diffgs}.

These approaches enable Gaussian-based generation through different
constraints on scene representation, including fixed global or local capacity,
occupancy-dependent support, and continuous functional parameterization.
\modelname{} instead places structural regularity in a fixed-layout triplane
latent. It directly encodes fitted Gaussian sets with scene-dependent
cardinality and density, jointly represents geometry and appearance, and
decodes native Gaussian primitives with adaptive spatial allocation and local
counts. This design provides a regular tensor interface for generative
modeling while retaining the adaptive organization of native 3DGS.

\noindent\textbf{Amodal reconstruction and generative 3D representations.}
Recovering complete 3D structure from partial visual observations is an
important setting for generative scene modeling. NOVA3R
\cite{chen2026nova3r} uses non-pixel-aligned scene tokens and a generative
point-cloud decoder to reconstruct visible and occluded geometry, while
VolFill \cite{ngo2026volfill} learns a volumetric generative prior for
single-view amodal scene reconstruction. GaussianAnything
\cite{lan2025gaussiananything} explores point-cloud-structured latent
generation for 3D objects. In contrast, TriFold targets a direct
Gaussian-to-latent-to-Gaussian interface for \emph{native scene-level 3DGS},
jointly representing geometry and appearance while recovering
content-adaptive primitive distributions and counts.

\section{Method}
\label{sec:method}

\begin{figure}
    \centering
    \includegraphics[width=1\linewidth]{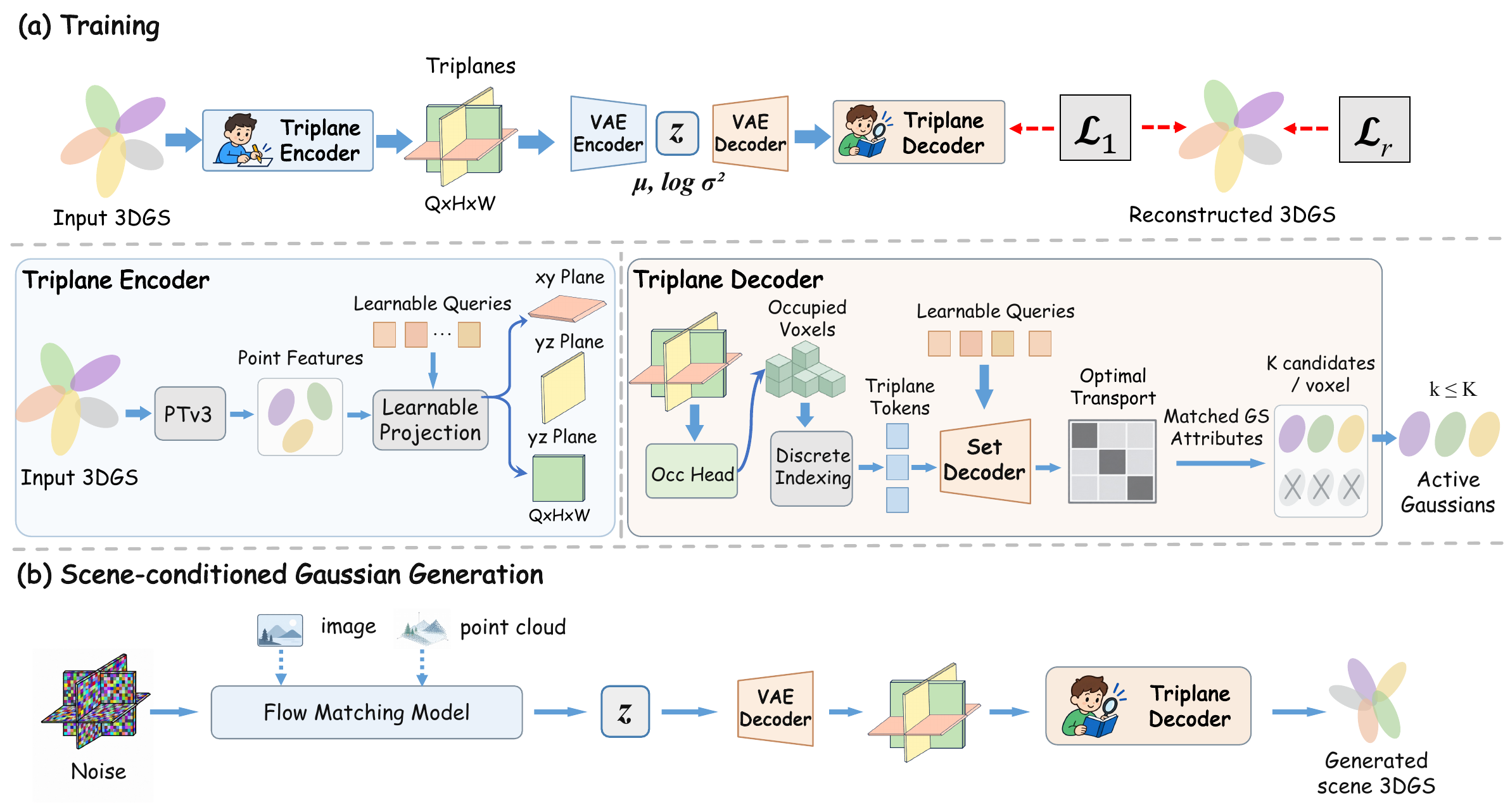}
    \vspace{-0.6cm}
    \caption{Overview of \modelname{}. The triplane encoder maps a variable-cardinality Gaussian set to a fixed-layout variational latent, and the triplane decoder reconstructs spatially adaptive local Gaussian sets. The learned latent further supports image-conditioned scene generation with flow matching.}
    \label{fig:overview}
\end{figure}

\subsection{Unified Geometry--Appearance Gaussian Latent Space}
\label{sec:overview}

3DGS scenes are unordered, variable-cardinality sets
 $\mathcal{G}=\{g_i\}_{i=1}^{N}$, where
$g_i=(x_i,{\ell}_i,q_i,\alpha_i,\mathbf{c}_i)$
denotes position, log-scale, rotation, opacity, and spherical-harmonic appearance. We learn a Gaussian-to-triplane VAE,
\begin{equation}
\mathbf{z}\sim q_{\phi}(\mathbf{z}\mid\mathcal{G}),
\qquad \widehat{\mathcal{G}}=D_{\psi}(\mathbf{z}),
\qquad
\mathbf{z}\in\mathbb{R}^{3C_z\times Q\times R_p\times R_p}.
\label{eq:vae_mapping}
\end{equation}
The difficulty is asymmetric: several spatially distinct Gaussians can project
to one plane cell, while one decoded cell can correspond to zero, one, or many
Gaussians. Our encoder therefore retains multiple features per projected cell,
and our decoder predicts a local Gaussian set only where the scene is occupied.
The fixed latent jointly represents geometry and appearance while the decoded
set remains spatially adaptive. We use $Q=4$, $C_z=16$, and $R_p=256$.

\subsection{Triplane Encoder}
\label{sec:encoder}

We use a modified PTv3 backbone~\citep{wu2024ptv3}, augmented with 3D rotary positional encoding, to extract a context-aware feature $\mathbf{h}_i$ for each Gaussian $g_i$.
For each plane $p\in\{xy,xz,yz\}$, we project the center of Gaussian onto the plane, assign the projection to grid cell $\mathbf{j}_i^p$, and concatenate $\mathbf{h}_i$ with its continuous projected coordinates, yielding the plane-conditioned feature $\widetilde{\mathbf{h}}_i^p$. For any cell ${\mathbf j}$, let
$\mathcal{I}_{{\mathbf j}}^p=\{i:{\mathbf j}_i^p={\mathbf j}\}$ denote the indices of Gaussians projected to cell \(\mathbf j\). Rather than pooling these features into a single descriptor, a query-based set aggregator produces $Q$ feature slots:
\begin{equation}
\mathbf{T}_{\mathbf{j}}^p=\operatorname{Agg}_{Q}
\bigl(\{\widetilde{\mathbf{h}}_i^p\}_{i\in\mathcal{I}_{\mathbf{j}}^p}\bigr)
\in\mathbb{R}^{Q\times C}.
\label{eq:writer_aggregation}
\end{equation}
Within the aggregator,  local geometric contextualization captures the relational topology among the projected features, after which $Q$ learned queries adaptively extract these features into discrete slots independently on each plane. 
The spatial projections of each primitive inherently induce geometric correspondences across the $xy$, $xz$, and $yz$ cells, enabling a sparse cross-plane attention module to propagate contextual information across corresponding slots prior to plane densification.
Finally, a convolutional variational
bottleneck predicts $(\boldsymbol{\mu},\boldsymbol{\sigma})$ and samples
$\mathbf{z}=\boldsymbol{\mu}+\boldsymbol{\sigma}\odot\boldsymbol{\epsilon}$.
This yields a regular tensor interface without collapsing multiple Gaussians projected to the same cell into a single feature. Architectural details are
provided in Appendix~\ref{sec:supp_vae_impl}.

\subsection{Triplane Decoder}
\label{sec:decoder}

The decoder reverses this mapping in two stages: support localization and Gaussian reconstruction.
It first samples and aggregates features from the three planes at each voxel center  $\mathbf{x}_v$, yielding context features $\mathbf{C}_v$, predicting occupancy
$\rho_v=f_{\mathrm{occ}}(\mathbf{C}_v,\mathbf{x}_v)$. Only voxels in
$\widehat{\mathcal{V}}=\{v\mid\rho_v>\tau_{\mathrm{occ}}\}$ are decoded during inference.
For each selected voxel, a DETR-style set decoder~\citep{carion2020detr}
uses $K$ learned queries conditioned on $\mathbf{C}_v$ to
predict Gaussian candidates $\{\widehat g_{vk}\}_{k=1}^{K}$ and their objectness probabilities $p^{\mathrm{obj}}_{vk}$. The queries are refined
through intra-voxel self-attention and sparse inter-voxel attention across neighboring voxels.
Because the local target is unordered and variable-sized, we use entropy-regularized Optimal Transport (OT) to softly match candidates to target Gaussians or a dummy ``no-Gaussian'' bin, with the matched target mass providing supervision for objectness.
At inference, candidates are retained according to their
objectness-opacity scores:
\begin{equation}
\widehat{\mathcal{G}}=
\{\widehat g_{vk}\mid v\in\widehat{\mathcal{V}},\;
p^{\mathrm{obj}}_{vk}\widehat\alpha_{vk}>\tau_{\mathrm{g}}\}.
\label{eq:reader_selection}
\end{equation}
Thus, \(K\) bounds only the local candidate capacity rather than fixing cardinality: both the spatial allocation of selected voxels and the number of retained primitives per voxel adapt to scene content, preserving a variable-cardinality native Gaussian representation.
Appendix~\ref{sec:supp_vae_impl} and ~\ref{sec:supp_ot_matching} provide the details of the architecture and OT.

\subsection{Joint Supervision for Gaussian VAE Training}
\label{sec:training}

We train the encoder and decoder to answer three questions:
\emph{where} Gaussian decoding should occur, \emph{what} local Gaussian set should be recovered, and \emph{which} predicted candidates should remain active at inference.

\paragraph{Occupancy supervision.}
Occupancy supervision should balance spatial coverage with sparsity. Positive-only supervision can lead to an unnecessarily dense active set, whereas treating uncertain boundary samples as hard negatives may suppress thin structures. We therefore combine pointwise occupancy regression with a voxel-center margin objective:
\begin{equation}
\mathcal{L}_{\mathrm{occ}}
=\lambda_{\mathrm{reg}}\,\mathbb{E}_{\mathbf{x}}
|\rho(\mathbf{x})-o(\mathbf{x})|
+\lambda_{\mathrm{m}}\!\left(
\mathbb{E}_{v\in\mathcal{V}_{+}}[m_{+}-\rho_v]_{+}^{2}
+\mathbb{E}_{v\in\mathcal{V}_{-}}[\rho_v-m_{-}]_{+}^{2}
\right).
\label{eq:occ_compact}
\end{equation}
Here $o(\mathbf{x})$ is the reference occupancy signal and \(\rho_v\) is the predicted occupancy at voxel center \(v\).
The sets $\mathcal{V}_{+}$ and $\mathcal{V}_{-}$ are occupied and confidently empty voxel centers, respectively. The margins \(m{+}\) and \(m_{-}\) encourage occupied centers to score above \(m_{+}\) and empty centers below \(m_{-}\).

\paragraph{Gaussian decoder supervision.}
For each occupied voxel \(v\), we use an unordered set of $M_v\leq K$ target Gaussians for set-level supervision, while the decoder predicts $K$ candidates. A detached OT plan $P_v^*$ softly matches the predicted candidates to these targets and defines the position loss
\begin{equation} \mathcal{L}_{\mathrm{pos}} =\frac{1}{|\mathcal{V}_{+}|} \sum_{v\in\mathcal{V}_{+}}\frac{K}{M_v} \sum_{k=1}^{K}\sum_{j=1}^{M_v} [P_v^*]_{kj}\, \frac{\|\widehat{\mathbf{x}}_{vk}-\mathbf{x}_{vj}\|_1}{h}, \label{eq:gs_decoder_compact} \end{equation}
where \(h\) denotes the voxel size. Under this correspondence, we directly supervise only Gaussian centers; scale, rotation, appearance, and opacity are instead constrained through rendering supervision. As in set-based object prediction, we use an objectness loss \(\mathcal{L}_{\mathrm{obj}}\) to predict whether each Gaussian corresponds to a target primitive, using the non-dummy transport mass as a soft target.

During rendering supervision, we render the selected Gaussian candidates from sampled cameras and compare them with target views rendered from the complete
ground truth Gaussian scene:
\begin{equation}
\mathcal{L}_{\mathrm{render}}
= (1-\eta)\|\widehat I^{r}-I^{r}\|_1
+\eta\bigl(1-\operatorname{SSIM}(\widehat I^{r},I^{r})\bigr).
\label{eq:render_compact}
\end{equation}
where $\eta$ balances the pixel-wise and structural similarity terms.
We stop gradients from $\mathcal{L}{\mathrm{render}}$ through the predicted Gaussian centers, so that it constrains scale, rotation, spherical-harmonic appearance, and opacity without directly optimizing their positions.
When a voxel contains more than \(K\) primitives, matched \(\ell_1\) supervision covers only a subset. Rendering against the full ground-truth scene instead encourages the decoder's \(K\) candidates to preserve the overall target appearance.
For occupied voxels, the decoder objective is
\begin{equation}
\mathcal{L}_{\mathrm{GS}}=
\lambda_{\mathrm{pos}}\mathcal{L}_{\mathrm{pos}}+
\lambda_{\mathrm{obj}}\mathcal{L}_{\mathrm{obj}}+
\lambda_{\mathrm{render}}\mathcal{L}_{\mathrm{render}}
\label{eq:gs_objective_compact}
\end{equation}
Appendix~\ref{sec:supp_ot_matching} provides further details on the complete
OT transport marginals, threshold-aligned activation objective, and rendering protocol.

\paragraph{Phantom-voxel supervision.}
Training the set decoder only on ground-truth occupied voxels introduces a discrepancy: at inference, false-positive occupancy proposals are also passed to the decoder and may produce ghost Gaussians. We therefore mine predicted
voxels that do not overlap ground-truth support as \emph{phantom voxels}. Since these voxels contain no target primitives, we apply
\begin{equation}
\mathcal{L}_{\mathrm{ph}}=
\mathbb{E}_{v\in\mathcal{V}_{\mathrm{ph}}}\!\left[
\frac{1}{K}\sum_{k=1}^{K}
\left(\lambda_{\mathrm{ph,obj}}\operatorname{BCE}
(p^{\mathrm{obj}}_{vk},0)
+\lambda_{\mathrm{ph},\alpha}|\widehat\alpha_{vk}|\right)
\right].
\label{eq:phantom_compact}
\end{equation}
These terms teach the decoder to reject proposer errors rather than assuming that
every decoded voxel contains a primitive. We exclude a small halo around
occupied voxels to avoid treating uncertain boundaries as negatives; sampling
details are given in Appendix~\ref{sec:supp_ot_matching}.

The KL term regularizes the latent for generative modeling. The complete
objective is
\begin{equation}
\mathcal{L}_{\mathrm{VAE}}=
\mathcal{L}_{\mathrm{occ}}+
\mathcal{L}_{\mathrm{GS}}+
\mathcal{L}_{\mathrm{ph}}+
\lambda_{\mathrm{KL}}\mathcal{L}_{\mathrm{KL}}.
\label{eq:vae_objective_compact}
\end{equation}
More detailed weights and the optimization schedule are reported in
Appendix~\ref{sec:supp_training}.

\subsection{Scene Generation over Gaussian Latents}
\label{sec:generation}

We freeze the VAE and train a conditional latent flow-matching model. For image
conditioning, dense DINO features are lifted with estimated depth and camera
geometry, splatted onto the three latent-aligned planes, and concatenated with
patch embeddings of the noisy scene latent. Given a normalized clean latent
$\mathbf{z}_0$, noise $\boldsymbol{\epsilon}$, and
$\sigma\in[0,1]$, we use the linear path
$\mathbf{z}_{\sigma}=(1-\sigma)\mathbf{z}_0+\sigma\boldsymbol{\epsilon}$ and
optimize
\begin{equation}
\mathcal{L}_{\mathrm{FM}}=
\mathbb{E}\!\left[
\frac{1}{D}\left\|v_{\theta}(\mathbf{z}_{\sigma},\sigma,\mathbf{c})-
(\boldsymbol{\epsilon}-\mathbf{z}_0)\right\|_2^2
\right].
\label{eq:flow_matching_compact}
\end{equation}
where $D$ denotes the dimensionality of the vectorized latent.
The same latent interface also supports point-cloud-conditioned completion.
Appendix~\ref{sec:supp_flow_impl} details how image and point-cloud
observations are converted into aligned condition tokens and injected into the flow transformer.

\begin{figure}[t]
    \centering
    \includegraphics[width=.98\linewidth]{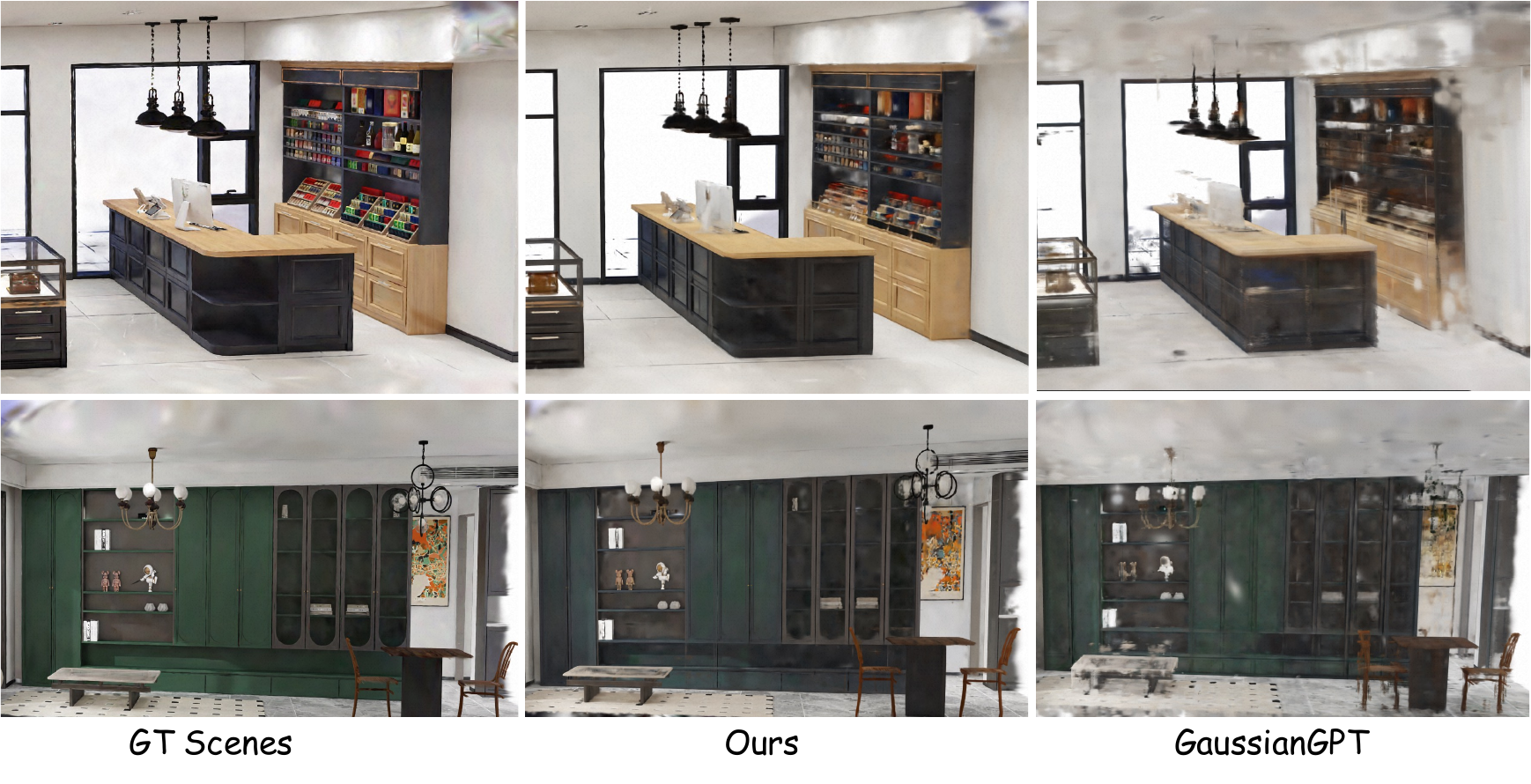}
    \vspace{-0.2cm}
\caption{\textbf{Qualitative VAE reconstruction comparison.} Compared with GaussianGPT, our method better captures scene geometry and appearance.}
    \label{fig:vae_comp}
    \vspace{-0.2cm}
\end{figure}

\section{Experiments}
\label{sec:experiments}

\subsection{Datasets and Evaluation Protocol}
\label{sec:datasets}

\paragraph{Datasets and View Sampling.}
We evaluate on InteriorGS~\citep{miao2026towards}, which contains large-scale structured indoor 3D Gaussian scenes, and the Aria Synthetic Environments (ASE) 3DGS subset of SceneSplat-49K~\citep{ma2025scenesplatpp}. Since InteriorGS scenes span large spatial regions, we crop them into smaller training samples following the procedure in Appendix~\ref{sec:supp_implementation}, yielding over 19,600 crops. Aria scenes are used directly without cropping, yielding approximately 8,600 samples. InteriorGS and Aria scenes are split 9:1 into training and test sets before cropping to avoid scene overlap. To construct single-image-to-amodal-3DGS pairs, we sample candidate conditioning views and use Qwen3-VL-8B~\cite{bai2025qwen3} to filter out uninformative views.

\begin{figure}[t]
    \centering
\includegraphics[width=0.98\linewidth]{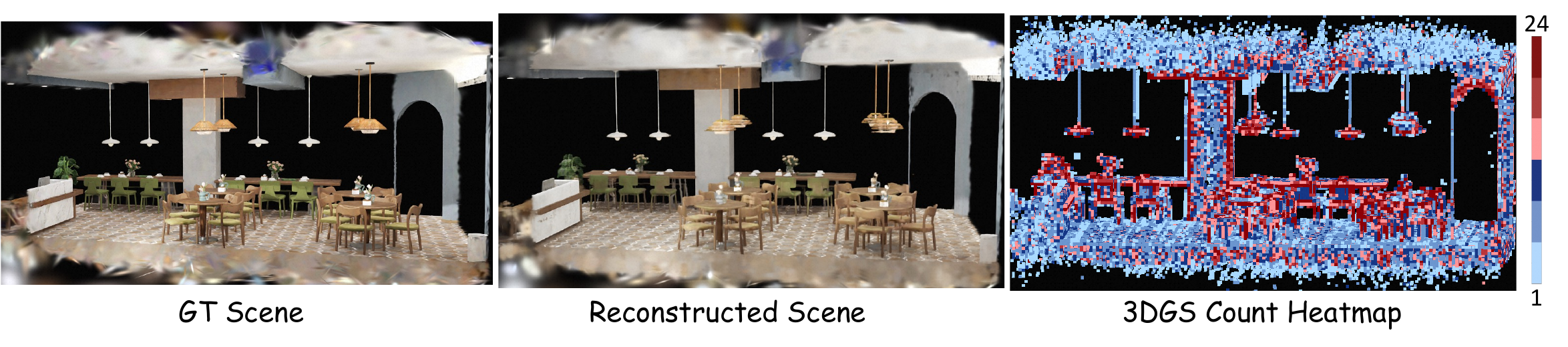}
    \vspace{-0.2cm}
    \caption{\textbf{Adaptive 3DGS allocation.} Visualization of the target scene, VAE reconstruction, and retained Gaussian count per occupied voxel. }
    \label{fig:adaptive_allocation}
        \vspace{-0.2cm}
\end{figure}

\begin{wraptable}{r}{0.5\linewidth}
  \centering
  \vspace{-0.5cm}
  \caption{Quantitative comparison with existing 3D Gaussian representation methods. Chamfer Distance (CD) values are multiplied by $10^3$. Best results are in \textbf{bold}, second-best are \underline{underlined}.}
    \label{tab:comparisons}
  \vspace{-0.2cm}
  \footnotesize
  \setlength{\tabcolsep}{4pt}
  \begin{tabular}{lcccc}
    \toprule
    Method & PSNR$\uparrow$ & SSIM$\uparrow$
           & LPIPS$\downarrow$ & CD$\downarrow$ \\
    \midrule
    Can3Tok
        & 4.53 & 0.012 & 0.920 & 10.96 \\
    DiffGS
        & 12.43 & 0.719 & 0.374 & 3.21 \\
    GaussianGPT
        & \underline{21.40} & \underline{0.856} & \underline{0.157} & \underline{2.62} \\
    \midrule
    \rowcolor{cyan!10}
    \textbf{\modelname{} (Ours)}
        & \textbf{26.55} & \textbf{0.913}
        & \textbf{0.088} & \textbf{2.23} \\
    \bottomrule
  \end{tabular}
  \vspace{-8pt}
\end{wraptable}
\paragraph{Implementation Details}
\label{sec:implementation}

The triplane encoder uses $Q=4$ tokens per plane cell, and the variational
latent has $C_z=16$ channels per plane at a spatial resolution of $256^2$.
The decoder predicts at most $K=24$ candidates per selected voxel on a
$256^3$ occupancy grid. We train the Gaussian VAE with Adam and a learning
rate of $2\times10^{-4}$. After convergence, we freeze the VAE, normalize its
latents with training-set statistics, and train the flow model with AdamW, learning rate of $10^{-4}$, and weight decay $0.01$. Both stages use
gradient clipping and cosine learning-rate schedules with linear warmup.
We train the VAE for 80K steps and the latent flow model for 100K steps on AMD MI308X GPUs, with a per-GPU batch size of 1. Complete data preprocessing and optimization settings are provided in
Appendices~\ref{sec:supp_implementation} and~\ref{sec:supp_training}.

%


\subsection{Scene Reconstruction}
\label{sec:reconstruction}
In this section, we evaluate how well the Gaussian VAE preserves scene appearance and geometry. We compare against Can3Tok~\citep{gao2025can3tok}, DiffGS~\citep{zhou2024diffgs}, and GaussianGPT~\citep{vonluetzow2026gaussiangpt}, which respectively use canonical tokens, continuous feature fields, and voxel-quantized representations. Table~\ref{tab:comparisons} reports rendering fidelity (PSNR, SSIM, and LPIPS) and geometric reconstruction (CD). Our method achieves the best results across all metrics, improving PSNR over GaussianGPT by 5.15 dB while reducing LPIPS and CD. Figure~\ref{fig:vae_comp} further shows that our VAE better preserves scene structure and appearance. Additional reconstructions are provided in Appendix Figure~\ref{fig:vae_comp_supp}, and Appendix~\ref{sec:latent_analysis} evaluates our VAE's latent smoothness and robustness. Using official implementations under the same scene-level training and evaluation protocol, adapted Can3Tok and DiffGS baselines converge to low-detail reconstructions, mapping distinct input scenes to similar coarse structures (Appendix Figure~\ref{fig:failure_case}).


\begin{table*}[t]
  \centering
    \caption{
    \textbf{Conditional generation evaluation.}
    For single-image conditioning, appearance is evaluated separately in regions observed and unobserved from the conditioning view.
    Full-point-cloud conditioning covers the complete scene and therefore has no unobserved split. Detailed evaluation protocols are provided
    in Appendix~\ref{supp:generation_metric}.
    }
  \vspace{-0.3cm}
  \label{tab:generation}
  \small
  \setlength{\tabcolsep}{5pt}
  \renewcommand{\arraystretch}{1.15}
  \resizebox{\textwidth}{!}{%
  \begin{tabular}{lcccccc}
    \toprule
    & \multicolumn{2}{c}{Observed Appearance}
    & \multicolumn{2}{c}{Unobserved Appearance}
    & \multicolumn{2}{c}{Unobserved Geometry} \\
    \cmidrule(lr){2-3}
    \cmidrule(lr){4-5}
    \cmidrule(lr){6-7}
    Condition
    & FID $\downarrow$
    & KID ($\times10^3$) $\downarrow$
    & FID $\downarrow$
    & KID ($\times10^3$) $\downarrow$
    & COV-CD (\%) $\uparrow$
    & MMD-CD ($\times10^3$) $\downarrow$ \\
    \midrule
    Full point cloud
    & 54.12 & 16.76
    & \multicolumn{4}{c}{N/A (complete conditioning)} \\
    Single image
    & 60.38 & 18.20
    & 67.81 & 21.43
    & 85.24 & 12.75 \\
    \bottomrule
  \end{tabular}}
  \vspace{-0.2cm}
\end{table*}

\begin{figure}[t]
    \centering
    \includegraphics[width=0.97\linewidth]{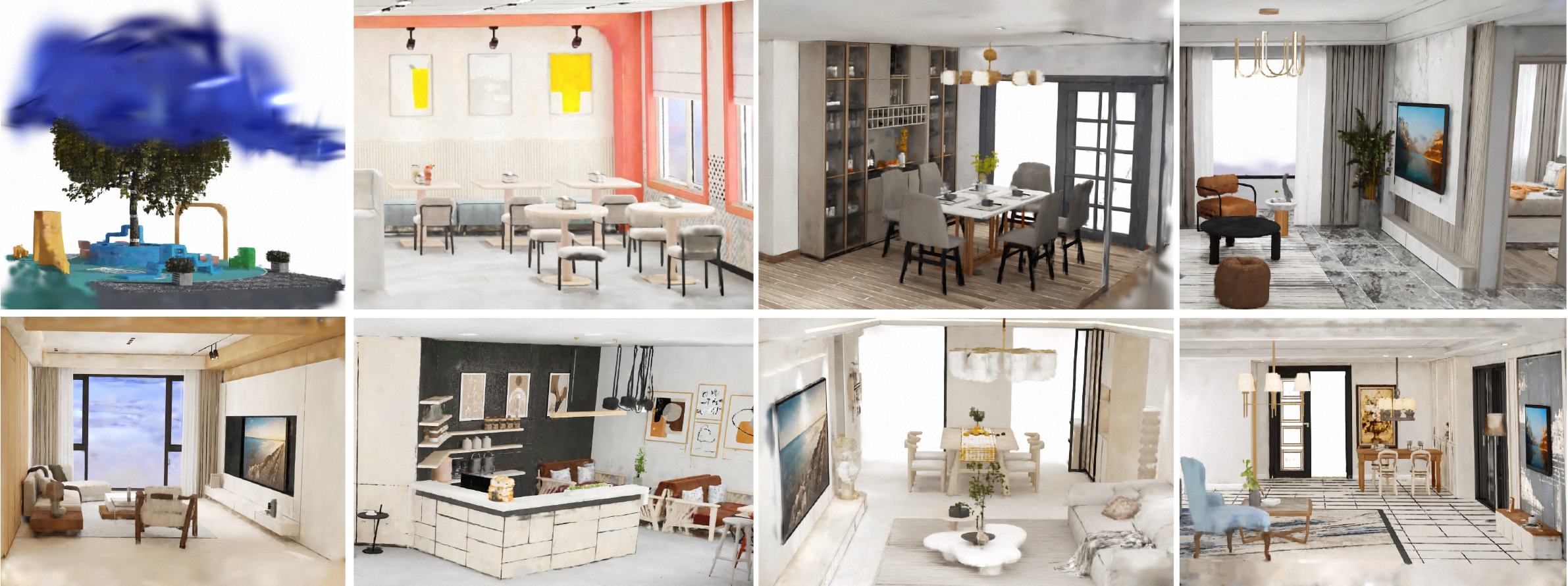}
    \vspace{-0.2cm}
    \caption{
    \textbf{Point-cloud-conditioned generation.}
Given a full point cloud, the latent flow model generates native 3DGS with
coherent geometry and appearance across novel views.
    }
    \vspace{-0.2cm}
    \label{fig:pcd_gen}
\end{figure}

\begin{figure}[ht]
    \centering
\includegraphics[width=0.97\linewidth]{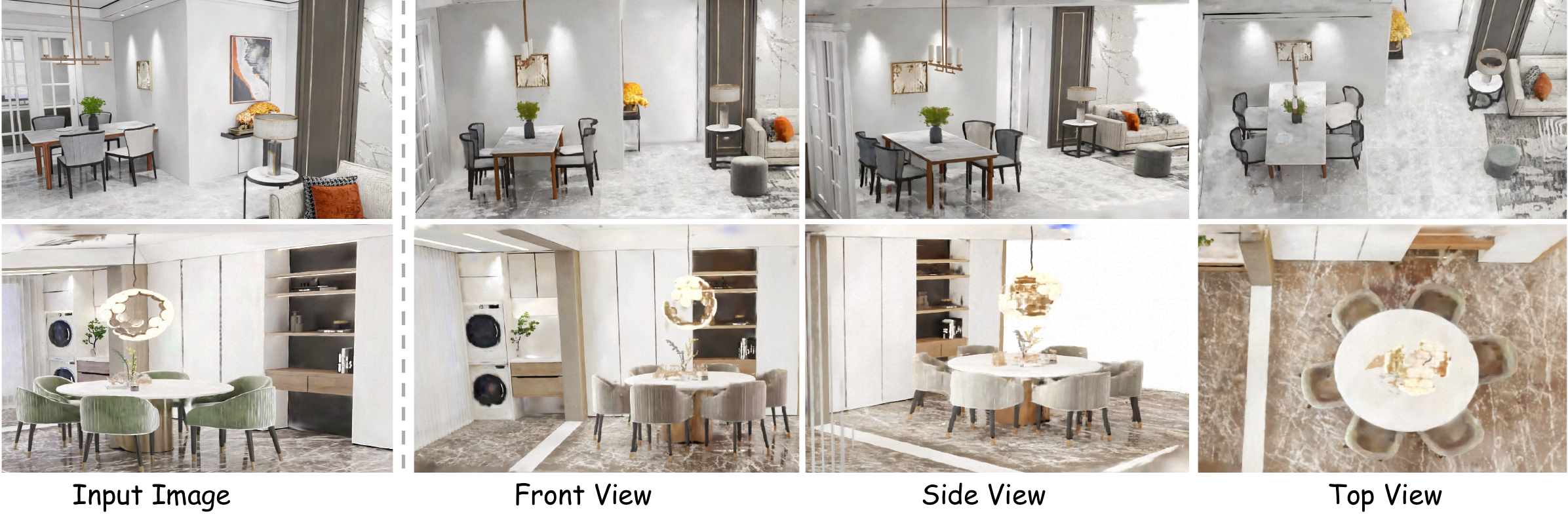}
    \vspace{-0.2cm}
\caption{
\textbf{Single-image amodal generation.}
Given one input image, the latent flow model generates plausible room
structure and appearance across novel views.
}
    \label{fig:image2gs_main}
    \vspace{-0.2cm}
\end{figure}

\paragraph{Adaptive Gaussian Allocation.}
Figure~\ref{fig:adaptive_allocation} compares the target and reconstruction
and shows the retained candidates per occupied voxel. The decoder predicts \(K=24\) candidate Gaussians per voxel and retains only those with sufficient objectness, allowing the number of output Gaussians to vary with local content. Thus, \(K\) specifies an upper bound rather than a fixed output count. We choose $K=24$ based on
the InteriorGS statistics in Figure~\ref{fig:stats_3dgs}, where 97.8\% of
occupied voxels contain at most 24 Gaussians on the same $256^3$ grid. For datasets with higher local densities, the decoder can be
configured with a larger $K$ without changing the latent layout.

\subsection{Scene Generation}
\label{sec:exp_generation}

To evaluate the generative utility of the learned \modelname{} latent, we
train latent flow-matching models under full-point-cloud and single-image
conditioning. Our latent supports generation from complete spatial
observations and amodal completion from a single view, while decoding to
content-adaptive native 3DGS. 
As existing 3DGS generators typically adopt different conditioning protocols, direct comparisons require method-specific adaptation; we focus on evaluating generation within our conditional setup rather than benchmarking across generative methods.
Qualitatively, Figures~\ref{fig:pcd_gen} and~\ref{fig:image2gs_main} show
plausible scene geometry and appearance across widely separated views.
In particular, the single-image model synthesizes plausible room structure and content
beyond the input view. Additional samples are provided in
Figures~\ref{fig:pcd2gs_supp} and~\ref{fig:image2gs_supp}.
Quantitatively, Table~\ref{tab:generation} shows better observed-region
appearance under full-point-cloud conditioning, while the single-image model
achieves an unobserved-region FID of 67.81 and a COV-CD of 85.24\%.
Since a single image only partially observes the scene, plausible completions
may differ from the reference in unobserved regions.

\subsection{Ablation Studies}
\label{sec:ablations}

\begin{table}[ht]
  \centering
  \setlength{\tabcolsep}{3pt}
  \begin{minipage}[t]{0.48\columnwidth}
    \centering
\caption{\textbf{Training data composition.} We compare models trained with different amounts of data and evaluate all variants on the same held-out test set.}
    \vspace{0.2cm}
    \label{tab:abl_data}
    \footnotesize
    \begin{tabular*}{\linewidth}{@{\extracolsep{\fill}}l|cccc@{}}
      \toprule
      Training data & PSNR$\uparrow$ & SSIM$\uparrow$ &LPIPS$\downarrow$  & CD$\downarrow$ \\
      \midrule
      + InteriorGS  & 25.83 & 0.904 & 0.096 & 2.27 \\
      + Aria          & \textbf{26.55} & \textbf{0.913} & \textbf{0.088} & \textbf{2.23} \\
      \bottomrule
    \end{tabular*}
  \end{minipage}
  \hfill
  \begin{minipage}[t]{0.48\columnwidth}
    \centering
    \caption{\textbf{Cross-dataset generalization.}
    Models are evaluated on the Aria test set \emph{without} any Aria training data;
    the last row is an in-domain reference trained with its training set.}
        \vspace{-0.2cm}
    \label{tab:abl_generalization}
    \footnotesize
    \begin{tabular*}{\linewidth}{@{\extracolsep{\fill}}l|cccc@{}}
      \toprule
      Training data & PSNR$\uparrow$ & SSIM$\uparrow$ &LPIPS$\downarrow$& CD$\downarrow$ \\
      \midrule
      InteriorGS     & 22.00 & 0.720 & 0.33 & 2.21 \\
      + Aria (in-domain) & \textbf{24.26} & \textbf{0.781} & \textbf{0.32} & \textbf{2.17} \\
      \bottomrule
    \end{tabular*}
  \end{minipage}
  \vspace{-0.2cm}
\end{table}

\begin{table*}[h]
\centering
\caption{\textbf{Component analysis.} Each row adds or replaces one model or training component while keeping the remaining evaluation protocol fixed. For the \(\mathcal{L}_{\mathrm{pos}}\)-only baseline, we use the same correspondence to directly supervise all Gaussian attributes.}
\vspace{-0.2cm}
\label{tab:ablation}
\small
\resizebox{\textwidth}{!}{%
\begin{tabular}{c|cc|ccc|ccc|c}
  \toprule
  \multirow{2}{*}{ID} & \multicolumn{2}{c|}{Architecture}
  & \multicolumn{3}{c|}{Training objective}
  & \multicolumn{3}{c|}{Rendering quality}
  & \multicolumn{1}{c}{Gaussian reconstruction} \\
  \cmidrule(lr){2-3}
  \cmidrule(lr){4-6}
  \cmidrule(lr){7-9}
  \cmidrule(lr){10-10}
  & Triplane Encoder & Triplane Decoder
  & Phantom & $\mathbf{\mathcal{L}_{\mathrm{pos}}}$ & $\mathcal{L}_{\mathrm{render}}$
  & PSNR$\uparrow$ & SSIM$\uparrow$
  & LPIPS$\downarrow$ & Chamfer Distance (CD) $\downarrow$  \\
  \midrule
   baseline &    &   &        & \cmark &        & 12.10 & 0.617  & 0.491  & 3.06 \\
  1        & \cmark &        &        & \cmark &        & 14.72 & 0.752 & 0.378 & 2.58 \\
  2        &        & \cmark &        & \cmark &        & 13.53 & 0.669 & 0.330 & 2.77 \\
  3        & \cmark & \cmark &        & \cmark &        & 17.81 & 0.803 & 0.231 & 2.48 \\
  4        & \cmark & \cmark &        &        & \cmark & 22.64 & 0.885 & 0.134 & 2.41 \\
  5        & \cmark & \cmark &        & \cmark & \cmark & 24.39 & 0.892 & 0.107 & 2.36 \\
  \midrule
    \rowcolor{cyan!10} Ours & \cmark & \cmark & \cmark & \cmark & \cmark & \textbf{25.83} & \textbf{0.904} & \textbf{0.096} & \textbf{2.27} \\
  \bottomrule
\end{tabular}%
}
\vspace{-0.2cm}
\end{table*}

\paragraph{Data Scaling and Cross-dataset Generalization.}
Unless otherwise specified, all ablations are trained on the InteriorGS training split. Under this setting, our model transfers to Aria without any Aria training data (Table~\ref{tab:abl_generalization}), demonstrating cross-dataset generalization. Separately, Table~\ref{tab:abl_data} shows consistent gains from scaling the training data, with further improvements when Aria is included.

\paragraph{Component Ablations.}
We next isolate the contribution of individual model components while keeping
the training data and evaluation protocol fixed.
Table~\ref{tab:ablation} analyzes the model in three stages. First, the
multi-query encoder and query-based set decoder each improve the baseline, while
combining them yields further gains, demonstrating their complementary roles in Gaussian encoding and decoding. Second, rendering supervision substantially improves visual fidelity, and combining it with the position loss further benefits both appearance and geometry. Finally, phantom-voxel training provides additional gains across all metrics, resulting in the full model.

\begin{wraptable}{r}{0.5\linewidth}
  \centering
  \vspace{-14pt}
    \caption{\textbf{Architectural design choices.}
    Comparison of alternative encoder, projection, and decoding designs under the same evaluation settings.}
  \label{tab:abl_writer_reader}
  \vspace{-0.2cm}
  \small
  \setlength{\tabcolsep}{3pt}
  \begin{tabular}{l|cccc}
    \toprule
    Variant & PSNR$\uparrow$ & SSIM$\uparrow$
            & LPIPS$\downarrow$ & CD$\downarrow$ \\
    \midrule
    \rowcolor{blue!5} \multicolumn{5}{l}{\emph{Encoder Head}} \\
    \quad Sparse 3D CNN & 23.16 & 0.830 &  0.139 & 2.48 \\
    \quad PointNet & 24.37 & 0.878 & 0.117 & 2.41 \\
    \quad \underline{PTv3} & \textbf{25.83} & \textbf{0.904} & \textbf{0.096} & \textbf{2.27} \\
    \midrule
    \rowcolor{blue!5} \multicolumn{5}{l}{\emph{Aggregation}} \\
    \quad Mean pooling & 24.60 & 0.857 & 0.102 & 2.39 \\
    \quad \underline{QFormer}  & \textbf{25.83} & \textbf{0.904} & \textbf{0.096} & \textbf{2.27} \\
    \midrule
    \rowcolor{blue!5} \multicolumn{5}{l}{\emph{Decoder Head}} \\
    \quad MLP & 23.09 & 0.885 & 0.132 & 2.55 \\
    \quad \underline{Set decoder} & \textbf{25.83} & \textbf{0.904} & \textbf{0.096} & \textbf{2.27} \\
    \bottomrule
  \end{tabular}
  \vspace{-10pt}
\end{wraptable}

\paragraph{Architectural Design Choices.}
Beyond the component-level ablation, Table~\ref{tab:abl_writer_reader} compares alternative encoder and decoder implementations with the rest of the pipeline fixed. For Gaussian encoding, point-based backbones outperform the sparse 3D CNN, with PTv3 performing best and capturing richer geometric information, as illustrated in Figure~\ref{fig:encoder_comparison}.
For aggregation, QFormer improves over mean pooling, supporting query-based
aggregation for resolving many-to-one collisions in triplane cells. For
decoding, the set decoder substantially outperforms the fixed-output MLP,
validating adaptive set prediction for recovering variable-cardinality
Gaussians. Together, these results show that the gains arise from architectural
choices tailored to the irregular and variable-density structure of 3DGS.

\section{Conclusion}
\label{sec:conclusion}

We introduced \textbf{\modelname{}}, a native scene-level 3D Gaussian VAE that maps Gaussian sets with adaptive cardinality and density into fixed-layout triplane latents. Rather than regularizing the Gaussian representation itself, the encoder captures fine-grained structure using cell-wise learned queries and primitive-anchored cross-plane attention, while the decoder predicts local Gaussian sets with scene-dependent allocations. The resulting latent jointly represents geometry and appearance and provides a regular interface for flow-based generative modeling. Experiments demonstrate high-fidelity reconstruction across diverse and complex scenes and show that the learned latent supports conditional scene generation, including amodal completion from partial observations.


\bibliographystyle{unsrt}
\bibliography{references}

\appendix
\section{Appendix}

In this appendix, we provide additional implementation and training details, along with extended experimental evaluations and qualitative results. These additions complement the main paper and further demonstrate the effectiveness and robustness of our approach.

\begin{figure}[h]
    \centering
\includegraphics[width=1\linewidth]{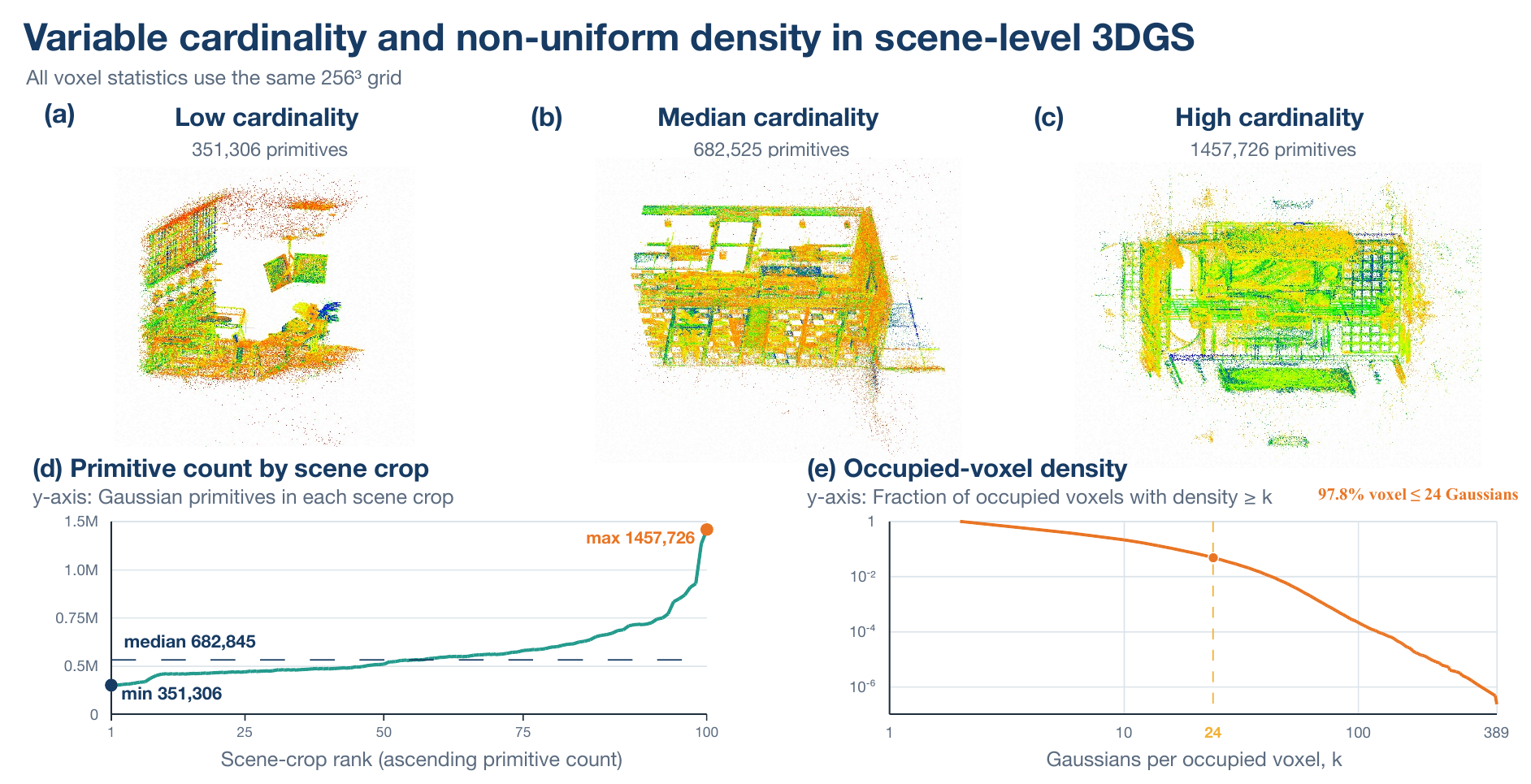}
\caption{
\textbf{Variable cardinality and nonuniform spatial density in scene-level 3DGS.}
(a-c) Representative InteriorGS scene crops with different primitive counts, visualized using Gaussian centers voxelized on a shared $256^3$ grid.
(d) Across 100 random scene crops, the number of primitives ranges from 351{,}306 to 1{,}457{,}726, with a median of 682{,}845.
(e) The complementary cumulative distribution of occupied-voxel density shows a pronounced long tail: 97.8\% of occupied voxels contain at most 24 Gaussians, while the maximum density can even reach 389 Gaussians per voxel.
These statistics motivate a representation that accommodates both scene-dependent cardinality and non-uniform local Gaussian allocation.
}
    \label{fig:stats_3dgs}
\end{figure}

\paragraph{Challenges of Native Scene-level 3DGS Modeling.}
As illustrated in Fig.~\ref{fig:stats_3dgs}, scene-level 3DGS exhibits
substantial variations in both global primitive cardinality and local spatial
density. Across scene crops, total primitive counts vary by more than a factor
of four, while the number of Gaussians within occupied voxels follows a
pronounced heavy-tailed distribution. Compounded by severe clutter and complex
occlusions inherent to indoor environments, these structural irregularities
are inherently mismatched with the fixed-dimensional tensor structures
required by standard latent generative models. A useful latent representation
should therefore regularize the representation layout while preserving
content-adaptive Gaussian allocation in the decoded scene.

By \emph{native 3DGS} modeling, we specifically refer to a
Gaussian-to-latent-to-Gaussian representation interface: our autoencoder
directly ingests pre-fitted Gaussian sets and reconstructs content-adaptive
Gaussian outputs, without converting the scene into an intermediate 3D
representation or entangling representation learning with 2D-to-3D visual
inference. This paradigm is fundamentally distinct from image-to-3DGS VAEs,
whose latent representations are jointly shaped by the image-conditioned 3D
reconstruction problem.

For computational tractability, our local set decoder uses $K=24$ candidate
queries per selected voxel. Here, $K$ defines the maximum local candidate
capacity rather than a fixed output cardinality: each selected voxel retains
a content-dependent subset of candidates, while the number and spatial
distribution of selected voxels remain scene-dependent. On the same
$256^3$ grid, $97.8\%$ of occupied voxels in the original fitted scenes
already contain at most 24 Gaussians. For the remaining high-density tail,
the set-level matching target retains at most $K$ primitives using an
opacity- and scale-aware selection strategy. Importantly, this local selection
does not limit the rendering target: rendering supervision is computed against
views rendered from the complete fitted Gaussian scene prior to local primitive
selection. The decoder is therefore supervised not only by local primitive
correspondence, but also by the collective rendering effect of the complete Gaussian representation.


\subsection{Data Preparation}
\label{sec:supp_implementation}

\paragraph{Gaussian Scene Normalization.}
Each input scene is represented as an unordered set of fitted Gaussian primitives. A primitive consists of 59 values: a 3D center, 48 spherical-harmonic coefficients for degree-3 appearance, one opacity logit, three anisotropic log-scales, and a four-dimensional rotation quaternion.

We use dataset-specific spatial normalization while maintaining a common canonical domain for the Gaussian VAE. For InteriorGS, each sample is constructed from a predefined 3D crop with a fixed spatial extent of \(B=8\). We translate the Gaussian centers by the corresponding crop center and apply the fixed transformation \(T_B\) that maps this crop to the canonical VAE domain. Because \(B\) is fixed, the spatial scale of this transformation is identical across all InteriorGS scenes. For Aria, which is used for Gaussian VAE evaluation but not for conditional diffusion training, no fixed \(B=8\) crop is applied; we instead determine a per-scene normalization scale from robust statistics of the Gaussian centers to map the valid scene extent into the same canonical domain.

In both cases, the model operates on a padded cubic domain of side length \(1.1\), discretized at a resolution of \(256^3\). The corresponding spatial transformation is applied consistently to Gaussian centers and occupancy-sample coordinates, and Gaussian log-scales are adjusted by the logarithm of its scale factor. Opacity, appearance, and rotation remain unchanged.

For set-level matching during Gaussian VAE training, an over-populated voxel retains at most 24 target primitives using an opacity- and scale-aware selection heuristic. This local selection is used only to construct the bounded set-level target; rendering supervision is computed from the complete fitted Gaussian scene before this selection.

\paragraph{Occupancy Targets Caching.}
The occupancy predictor is supervised by sampled 3D locations rather than by dense enumeration of all $256^3$ voxels. For each scene, we construct a nearest-center distance field from the preprocessed Gaussian centers and draw samples across four distinct regions: exact Gaussian centers, a thin inner shell, a near-surface transition shell, and far-field points. Center and inner-shell samples represent occupied space, the transition shell decays linearly toward zero, and far-field samples serve as negatives. In our experimental configuration, the two shell radii correspond to 0.47 and 1.19 voxels at resolution 256, and the four groups account for 15\%, 20\%, 35\%, and 30\% of the cached samples, respectively. Samples outside the padded model domain are removed after applying the corresponding scene scale. At each training iteration, we randomly sample 250{,}000 cached locations for occupancy supervision. This hierarchical sampling strategy sharpens the decision boundary while maintaining sufficient coverage of empty space.

\subsection{Gaussian VAE Architecture}
\label{sec:supp_vae_impl}


\paragraph{Multi-slot Writing and Cross-plane Aggregation.}
The point features are orthographically projected onto the $xz$, $xy$, and $yz$ planes at a resolution of $R_p=256$. Within each occupied cell, a four-layer, eight-head QFormer employs $Q=4$ learned queries to read the unordered set of projected primitives. Prior to attention, each projected feature is encoded as:
\begin{equation}
\widetilde{\mathbf{h}}_i^p=\operatorname{MLP}\!\left(
[\,\mathbf{h}_i+\mathbf{e}_p,\;\gamma(\mathbf{u}_i^p),\;
\gamma(\boldsymbol{\delta}_i^p)\,]\right),
\qquad
\boldsymbol{\delta}_i^p=R_p\mathbf{u}_i^p-
\lfloor R_p\mathbf{u}_i^p\rfloor,
\label{eq:supp_writer_position}
\end{equation}
where $\mathbf{u}_i^p\in[0,1)^2$ is the normalized in-plane coordinate, $\boldsymbol{\delta}_i^p$ is its within-cell offset, $\mathbf{e}_p$ is a learned plane embedding, and $\gamma$ represents an eight-frequency Fourier encoding. The learned queries also receive the Fourier encoding of the cell center. Collectively, these terms encode the specific plane, cell, and location of the primitive, while the PTv3 feature $\mathbf{h}_i$ provides the necessary 3D context. The multiple queries capture diverse summaries of the primitives colliding within a cell, preventing the information bottleneck caused by single-vector pooling. 

Subsequently, the encoder performs a single sparse cross-plane aggregation step. For each primitive, its three projected cells constitute an aligned triplet; tokens sharing the same query index attend across this triplet, and updates common to multiple primitives are averaged. This mechanism allows the orthogonal planes to reconcile complementary views of the same 3D geometry without requiring dense attention across all plane cells. Finally, the VAE maps each plane to $C_z=16$ variational channels using a base width of 384, channel multipliers $[1,2]$, and three residual blocks per resolution. Stacking the three planes along the channel dimension yields a $48\times4\times256\times256$ latent tensor. 

\paragraph{Occupancy-guided Support Proposal.}
At each candidate voxel center $\mathbf{x}_v$, the predictor retrieves $Q$ features from each plane using the exact same integer cell indexing as the Gaussian decoder. For every slot $q$, it concatenates the corresponding $xz$, $xy$, and $yz$ features and projects the resulting $3C$-dimensional vector back down to $C$ dimensions. The predictor then flattens the $Q$ fused features, appends $\mathbf{x}_v$ alongside its six-frequency Fourier encoding, and processes the result through a skip-connected MLP containing two four-layer blocks. A sigmoid activation maps the output logit to the occupancy probability $\rho_v$. During inference, we evaluate the $256^3$ voxel centers in chunks and forward cells with $\rho_v>0.4$ to the set decoder. This shared indexing guarantees that the occupancy predictor and Gaussian decoder consistently access the identical three plane cells at any given voxel.

\paragraph{DETR-style set decoder.}
Since the decoder must decode a regular triplane into an irregular point set of unknown local cardinality, we formulate the decoding process as a set prediction problem rather than fixed-index regression. First, an occupancy predictor identifies candidate voxels. For each candidate, 24 learned queries cross-attend to the three sampled plane contexts to explain the local Gaussian set. Separate task heads predict the primitive's offset, scale, rotation, spherical-harmonic appearance, opacity, and objectness. Sinkhorn matching establishes an order-independent correspondence to the target set (Sec.~\ref{sec:supp_ot_matching}); unmatched queries are routed to a dummy target. At inference time, inactive candidates are dynamically pruned based on their predicted objectness and opacity.

\subsection{Objectness-Aware Set Prediction and OT Matching}
\label{sec:supp_ot_matching}

\paragraph{Objectness and Differentiable Selection Supervision.}
Objectness indicates the probability that a query should successfully instantiate a Gaussian, whereas opacity defines that instantiated Gaussian's rendering contribution. Decoupling these roles ensures that unused query slots aren't penalized solely via near-zero opacity. The soft objectness target is derived from the non-dummy mass assigned by the Sinkhorn algorithm and is supervised via a balanced binary cross-entropy loss. During rendering and inference, query $k$ is ranked by a unified score:
\begin{equation}
s_k=p^{\mathrm{obj}}_k\,\alpha_k,
\label{eq:supp_query_score}
\end{equation}
where $\alpha_k$ represents its predicted opacity. In our reported VAE configuration, the selection threshold is set to 0.1. A differentiable activation loss forces at least one matched query in every occupied voxel to surpass a margin of 0.15, while suppressing unmatched queries below the selection boundary. During training, the renderer utilizes the union of OT-matched queries and those queries whose predicted scores naturally exceed the threshold. Consequently, a false positive selected by the model is directly exposed to the rendering objective, rather than being shielded by the target assignment.

Because candidate enumeration inherently involves a discrete thresholding step, we do not differentiate through the set membership operation itself. Instead, the occupancy, objectness, opacity, and activation scores governing this set are jointly optimized using differentiable surrogate losses aligned with the inference rule. As a result, the threshold functions as a deployment decision based on learned scores, rather than a manually tuned, post-hoc pruning heuristic.

\paragraph{Optimal-transport matching.}
Within each occupied voxel $v$ (of size $h$), the decoder predicts a fixed set of $K$ candidate Gaussians to reconstruct a variable-sized, unordered target set of $M_v \leq K$ Gaussians. To absorb candidate slots that do not correspond to any target primitive, we append an additional dummy ``no-Gaussian'' entry. We define the pairwise matching cost solely based on Gaussian centers:
\begin{equation}
[\mathbf{A}_v]_{kj} =
\begin{cases}
\frac{1}{h}\,\lVert \widehat{\mathbf{x}}_{vk} - \mathbf{x}_{vj} \rVert_1, & 1 \leq j \leq M_v,\\
0, & j = M_v + 1,
\end{cases}
\label{eq:supp_ot_cost}
\end{equation}
where $\widehat{\mathbf{x}}_{vk}$ and $\mathbf{x}_{vj}$ denote the centers of the $k$-th candidate and the $j$-th target Gaussian in voxel $v$, respectively. We obtain a soft correspondence by solving an entropy-regularized optimal transport problem:
\begin{equation}
P_v^{*} = \arg\min_{P \in \Pi(\mathbf{a}_v, \mathbf{b}_v)} \;
\langle \mathbf{A}_v, P \rangle_F - \varepsilon\, \mathcal{H}(P),
\label{eq:supp_ot}
\end{equation}
where $\mathcal{H}(P) = -\sum_{k,j} P_{kj} \log P_{kj}$ and $\Pi(\mathbf{a}_v, \mathbf{b}_v)$ denotes the set of transport plans subject to the marginals
\begin{equation}
\mathbf{a}_v = \frac{1}{K}\mathbf{1}_K,
\qquad
\mathbf{b}_v =
\left[
\frac{1}{K}\mathbf{1}_{M_v}^{\top},\;
1 - \frac{M_v}{K}
\right]^{\top}.
\label{eq:supp_ot_marginals}
\end{equation}
By construction, each target Gaussian receives a mass of $1/K$, while the dummy entry absorbs the residual mass $1 - M_v/K$ from unmatched candidate slots. We solve for $P_v^{*}$ with Sinkhorn iterations and stop gradients through it, i.e., the transport plan is treated as constant during backpropagation.

Under this correspondence, we directly supervise only Gaussian centers; scale, rotation, appearance, and opacity are instead constrained through rendering supervision. The same transport plan also provides a soft objectness target. Since each candidate has a row mass of $1/K$, its total non-dummy transport mass yields
\begin{equation}
t_{vk} = K \sum_{j=1}^{M_v} [P_v^{*}]_{kj}
= 1 - K\,[P_v^{*}]_{k, M_v + 1}.
\label{eq:supp_objectness_target}
\end{equation}
Candidates assigned to real Gaussians approach a target of $1$, whereas those routed to the dummy entry approach $0$. We optimize the corresponding objectness logits with a binary cross-entropy loss, balanced by the number of positive and negative candidates within each batch. Thus, a single center-based OT correspondence serves two purposes: it assigns each candidate a target Gaussian for center supervision, and it indicates whether the candidate represents a valid primitive, eliminating any separate matching heuristic for objectness.

\paragraph{Phantom Voxels Regularization.}
If the decoder were trained exclusively on ground-truth occupied voxels, it would never encounter false positives, even though the learned occupancy proposer will inevitably produce them during inference. This train-test domain gap would cause queries at erroneous locations to decode ``ghost'' Gaussians with high objectness or opacity. We bridge this gap by deliberately feeding selected false-positive proposals to the decoder as \emph{phantom voxels}. Because they contain no target Gaussians, all queries within a phantom voxel receive negative objectness supervision, forcing their activation and opacity to be heavily suppressed. Phantom training commences after the initial occupancy warm-up phase; we subsequently alternate between ground-truth and predicted occupancy supports, enabling the decoder to master both accurate local reconstruction and robust recovery from proposer errors. Although sampling the phantom set is non-differentiable, all losses evaluated on the selected queries are strictly differentiable with respect to the decoder and its prediction heads. This mechanism operates analogously to hard-negative mining, directly optimizing the network against failure cases it will realistically face at test time.

\subsection{Baseline Implementation and Training}
\label{sec:baseline_impl}

For Can3Tok, DiffGS, and GaussianGPT, we use the official implementations released by the respective authors. To ensure a controlled comparison, we adapt only their data-loading pipelines as necessary to train and evaluate all methods on the same scene-level 3DGS dataset and train/test split used by TriFold. Unless otherwise stated, we retain the original model architectures, representation designs, training objectives, and method-specific optimization settings, including the optimizer, learning-rate schedule, batch size, and loss weights, from the corresponding official training configurations.

Can3Tok's official implementation requires a fixed primitive budget of 40{,}000 Gaussians per scene. We therefore downsample each pre-fitted 3DGS scene to 40{,}000 Gaussians before tokenization and keep this budget fixed across all scenes. All baseline models are trained from scratch for 80K optimization steps, using the same number of optimization steps as our Gaussian VAE. Evaluation is performed on the same held-out scenes using the same rendering views and evaluation protocol for all methods.

Under this setting, GaussianGPT converges to meaningful scene-level reconstructions, whereas the adapted Can3Tok and DiffGS models converge to substantially lower-fidelity solutions. In particular, distinct input scenes are frequently reconstructed as similar low-detail Gaussian structures, as illustrated in Figure~\ref{fig:failure_case}. We report the resulting models directly rather than excluding these degraded cases from the comparison. These results characterize the behavior of the released methods under our scene-level adaptation and evaluation setting and should not be interpreted as general limitations of the original methods in the domains for which they were designed.

\paragraph{Geometric evaluation of degenerate predictions.}
We evaluate center-wise Chamfer Distance (CD) for all methods using the same preprocessing and the common normalized scene coordinate frame, without method-specific post-hoc alignment. Under our scene-level adaptation, the lower-fidelity baseline solutions can exhibit degenerate Gaussian attributes and substantial drift of the predicted centers from the reference geometry. We retain these errors in the geometric evaluation rather than realigning the predictions or excluding degraded scenes, such that CD directly reflects the resulting geometric discrepancy.

Accordingly, the large CD values of the affected baselines reflect their substantial geometric degradation rather than differences in the evaluation protocol. Rendering-based metrics (PSNR/SSIM/LPIPS) are evaluated independently using the same rendering views for all methods. Thus, both geometric and rendering metrics are computed under a common evaluation protocol, and degraded baseline predictions are reported directly rather than removed or assigned special-case values.

\subsection{Conditional Latent Flow Matching}
\label{sec:supp_flow_impl}


\paragraph{Condition Injection.}
The image-to-GS and point-cloud-to-GS models share a unified context-conditioning architecture. Each input observation is first converted into a dense spatial representation aligned with the Gaussian VAE latent grid. These representations are then embedded as condition tokens and concatenated with the noisy scene tokens, and the combined sequence is processed via joint self-attention. To distinguish their roles, condition tokens are modulated using the clean timestep ($\sigma=0$) via AdaLN, whereas scene tokens are modulated by the sampled noise level. Both token sets share the same two-dimensional RoPE coordinates at corresponding spatial locations but use separate patch embeddings to distinguish their modalities.

\paragraph{Image-condition preprocessing and coordinate alignment.}
Our conditional diffusion model is trained only on InteriorGS; Aria is used for evaluating the Gaussian VAE and is not included in diffusion training. Accordingly, the fixed $B=8$ spatial extent described below applies specifically to InteriorGS.

For image-to-GS conditioning, we resize each conditioning image to $720\times1280$ and extract dense DINOv3 ViT-H+/16 features~\cite{simeoni2025dinov3}, yielding a $45\times80$ patch grid with 1280 feature channels. We additionally estimate monocular depth using Depth Anything v3~\cite{lin2025depth}. Since monocular depth is defined only up to a global scale, we rescale the raw prediction $\widetilde d_n$ as
\begin{equation}
d_n = \gamma_d \widetilde d_n,
\label{eq:supp_depth_scale}
\end{equation}
where $\gamma_d$ is a single global depth-scale factor shared across all InteriorGS scenes and fixed throughout training and inference. This factor resolves the global scale ambiguity of monocular depth and is independent of the target 3DGS.

We discard the lowest-confidence $10\%$ of valid depth pixels and back-project the remaining DINO features using the corresponding camera intrinsics and extrinsics. For a valid sample $n$ with homogeneous image coordinate $\widetilde{\mathbf{u}}_n$, the back-projected 3D location is
\begin{equation}
\mathbf{x}^{\mathrm{world}}_n =
R^{\top}
\left(
d_n K^{-1}\widetilde{\mathbf{u}}_n - t
\right),
\label{eq:supp_dino_backprojection}
\end{equation}
where $(R,t)$ denotes the world-to-camera transformation. After applying $\gamma_d$, the predicted depth and camera translation are expressed in the same metric units.

All InteriorGS samples used for conditional diffusion training and evaluation are constructed from predefined 3D crops with a fixed spatial extent of $B=8$. Let $\mathbf{c}$ denote the center of the corresponding crop. We first express each lifted point in crop-local coordinates,
\begin{equation}
\mathbf{x}^{\mathrm{crop}}_n =
\mathbf{x}^{\mathrm{world}}_n-\mathbf{c},
\label{eq:supp_crop_local}
\end{equation}
and then apply the fixed crop-to-canonical transformation
\begin{equation}
\mathbf{x}^{\mathrm{tri}}_n =
T_B\left(\mathbf{x}^{\mathrm{crop}}_n\right).
\label{eq:supp_condition_normalization}
\end{equation}
Because the crop extent $B$ is fixed, $T_B$ uses the same spatial scale for every InteriorGS scene; only the crop translation $\mathbf{c}$ varies with the predefined crop location. Crucially, the same crop translation and transformation $T_B$ are applied to the Gaussian scene before encoding it into the target VAE latent. The lifted image features and target Gaussian latent therefore share the same canonical coordinate frame.

Both $\mathbf{c}$ and $B$ are specified by the InteriorGS data construction and are available independently of the target Gaussian representation. Constructing the image condition at inference therefore requires only the conditioning RGB image, its predicted depth, the associated camera parameters, the fixed depth-scale factor $\gamma_d$, and the predefined crop transform. No target 3DGS coordinates or target-derived spatial statistics are used.

We orthographically project each lifted DINO feature $\mathbf{f}_n$ at $\mathbf{x}^{\mathrm{tri}}_n$ onto the $xz$, $xy$, and $yz$ planes. For plane $p$, let $\mathcal{P}_{\mathbf{j}}^p$ denote the set of lifted samples whose projection falls into cell $\mathbf{j}$. Their features are aggregated by mean splatting:
\begin{equation}
\mathbf{S}_{\mathbf{j}}^p =
\frac{1}{|\mathcal{P}_{\mathbf{j}}^p|}
\sum_{n\in\mathcal{P}_{\mathbf{j}}^p}
\mathbf{f}_n,
\label{eq:supp_dino_splat}
\end{equation}
with empty cells zero-filled. This produces three condition planes spatially aligned with the target Gaussian triplane latent. We concatenate the three planes along the channel dimension and apply a dedicated $2\times2$ convolutional patch embedding to obtain $128^2=16{,}384$ spatial condition tokens.

\paragraph{Point-cloud conditioning.}
For point-cloud-to-GS conditioning, we apply the same spatial preprocessing used to construct the target Gaussian VAE latent. We retain only the Gaussian positions and colors, while replacing opacity, scale, and rotation with fixed neutral values before passing the resulting set through the frozen Stage-1 triplane encoder and VAE encoder. Specifically, we use an opacity logit of $40$, an isotropic log-scale of $\log(0.07h)$ (with voxel size $h=1.1/256$), and an identity quaternion.

The resulting condition latent has shape $48\times4\times256\times256$. We standardize it using the same training-set statistics as for the target VAE latents, flatten the four slots and 48 channels into a 192-channel tensor, and apply a separate $2\times2$ patch embedding to obtain $16{,}384$ spatially aligned condition tokens. For both conditioning modes, the complete condition is replaced by zeros with probability $10\%$ during training for classifier-free guidance.

\begin{figure}[t]
    \centering
    \includegraphics[width=\linewidth]{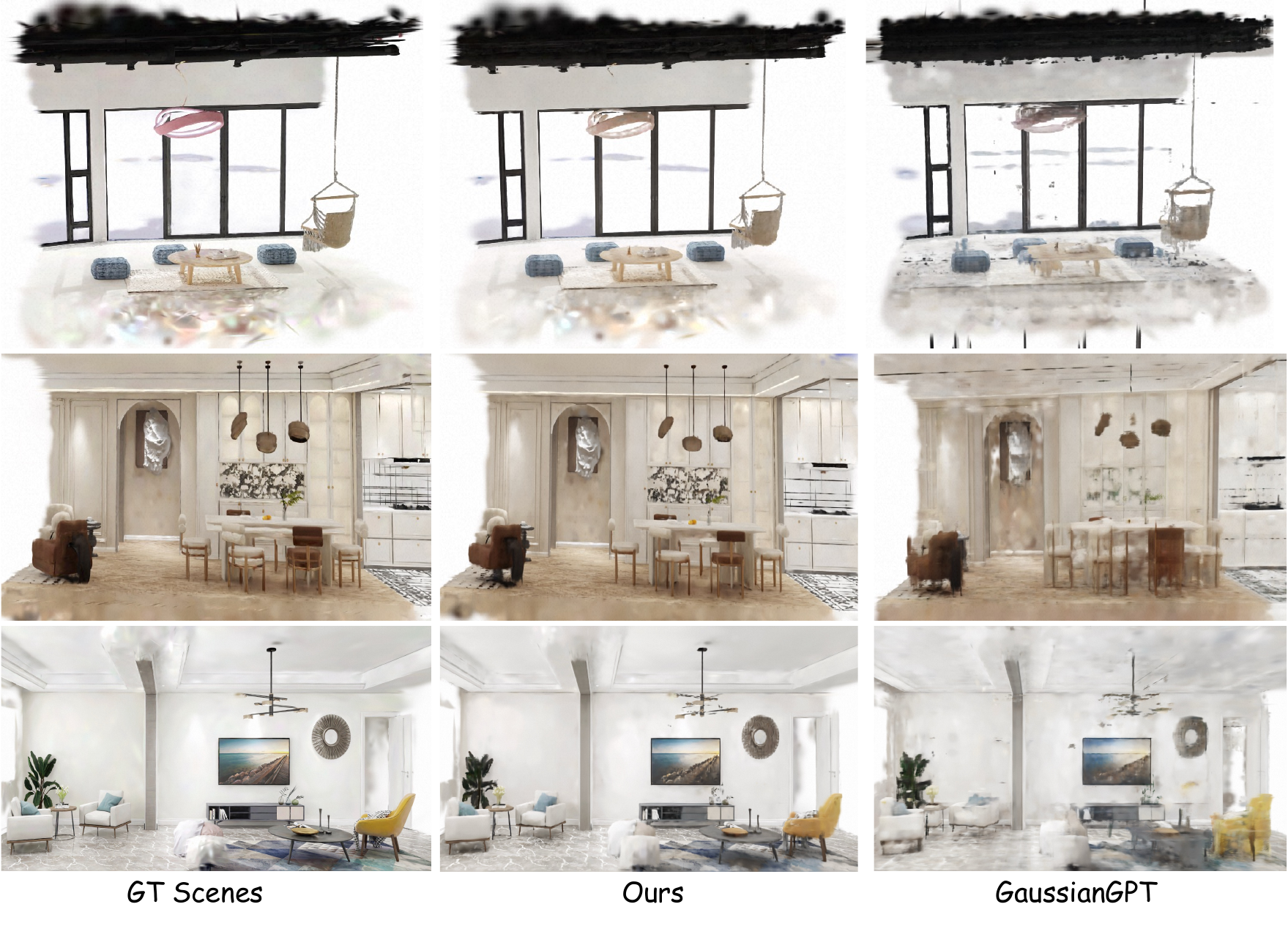}
    \caption{\textbf{Additional VAE Reconstructions.} Under the identical evaluation setup, our VAE preserves more visible layout, appearance, and fine-grained local structures in these examples.}
    \label{fig:vae_comp_supp}
\end{figure}

\noindent\textbf{Relation to Geometry-Aware Visual Conditioning.}
Constructing spatially aligned conditioning signals from
images is closely related to geometric perception and
correspondence estimation. Prior stereo methods have
explored local surface structures through slanted-plane
aggregation \cite{wang2022spnet}, disparity-aware cost
aggregation \cite{wang2024dcanet}, and structure-preserving
adaptive downsampling and disparity alignment
\cite{wang2025adstereo}. Recent approaches further exploit
visual foundation models to learn geometry-sensitive
and robust representations
\cite{zhang2024formerstereo,wang2025smoestereo,wang2026smformer},
while self-supervised methods improve geometric supervision
through constrained pseudo-label generation
\cite{wang2025dualnet}. While these approaches focus on
stereo correspondence estimation, TriFold employs pretrained
monocular depth and frozen visual features to construct
spatially aligned triplane conditions. Our primary contribution
lies in learning a structured generative representation of
native scene-level 3D Gaussians, rather than developing a
new geometric estimator.

\subsection{Training and Evaluation Details}
\label{sec:supp_training}

\paragraph{Gaussian VAE training.}
We jointly train the PTv3 encoder, the variational triplane autoencoder, the occupancy predictor, and the local Gaussian set decoder end to end. The latent representation comprises four slots, each with 16 channels per plane, at a spatial resolution of $256^2$. The decoder predicts $K=24$ candidate Gaussians per selected voxel, with the active subset determined by predicted objectness.
The overall objective combines occupancy, center reconstruction, objectness, rendering, and phantom-voxel suppression losses, together with KL regularization. As detailed in the main paper, optimal-transport matching operates solely on Gaussian centers and provides both the correspondence for center supervision and the soft objectness targets. Among the Gaussian attributes, only the normalized center coordinates are directly supervised under this correspondence, with a loss weight of 5. Scale, rotation, appearance, and opacity receive no direct matched-attribute supervision and are instead constrained through differentiable rendering, with additional opacity suppression on phantom voxels.

For the remaining objectives, the point-sampled and voxel-level occupancy losses use weights of 1 and 0.5, respectively; the objectness BCE and threshold-aligned activation losses use weights of 1 and 0.3; and the phantom-voxel objectness and opacity suppression terms use weights of 0.5 and 0.2. Direct spherical-harmonic regression is disabled in the reported configuration. We train the network with Adam at a peak learning rate of $2\times10^{-4}$. The KL weight is linearly warmed up to $2\times10^{-4}$ over the first 6k steps, and gradient norms are clipped to 1.

For completeness, we describe the foreground-aware RGB loss used for rendering supervision. Rendered training views often contain a large black background, so a uniform full-image average would disproportionately weight background pixels. We therefore compute the RGB $L_1$ error separately over foreground and background regions and downweight the latter. Pixels whose target alpha exceeds $0.05$ are treated as foreground. Let $\mathcal{F}^{r}$ and $\mathcal{B}^{r}$ denote the foreground and background pixel sets in view $r$, respectively, and let
\begin{equation}
\ell_{\mathcal{S}}^{r}
=
\frac{1}{|\mathcal{S}|}
\sum_{p\in\mathcal{S}}
\left\lVert \widehat{I}_{p}^{r}-I_{p}^{r} \right\rVert_{1}
\label{eq:supp_render_set_l1}
\end{equation}
denote the mean RGB $L_1$ error over a pixel set $\mathcal{S}$. The RGB loss in our implementation is then
\begin{equation}
\ell_{\mathrm{rgb}}^{r}
=
\frac{
w_{\mathrm{fg}}\,\ell_{\mathcal{F}}^{r}
+
w_{\mathrm{bg}}\,\ell_{\mathcal{B}}^{r}
}{
w_{\mathrm{fg}}+w_{\mathrm{bg}}
},
\qquad
w_{\mathrm{fg}}=1,\quad
w_{\mathrm{bg}}=0.1.
\label{eq:supp_render_weighted_l1}
\end{equation}
Thus, the RGB term $\mathcal{L}_{1}$ in the main paper is implemented as this foreground-aware $L_1$ loss rather than a uniform full-image average. The complete rendering objective is
\begin{equation}
\mathcal{L}_{\mathrm{render}}
=
\mathbb{E}_{r}\left[
(1-\eta)\,\ell_{\mathrm{rgb}}^{r}
+
\eta\bigl(1-\operatorname{SSIM}(\widehat{I}^{r},I^{r})\bigr)
\right],
\qquad
\eta=0.2.
\label{eq:supp_render_loss}
\end{equation}

At each training iteration, we randomly sample 32 camera viewpoints per scene and render the full decoded Gaussian set at $640\times640$ resolution, comparing the renderings against the corresponding target images. For rendering supervision, all decoded candidates are rendered before objectness-based primitive selection, providing scene-level supervision beyond the capacity-limited center targets used for set-level matching.

\paragraph{Gaussian Generation Training.}
We train a Wan-style latent transformer~\cite{wan2025wan} consisting of 30 blocks, a hidden width of 1536, a feed-forward width of 8960, and 12 attention heads. The four-slot latent is patchified using non-overlapping $2\times2$ spatial patches, yielding $128\cdot128=16{,}384$ scene tokens. The DINO-splat tokens are concatenated with the noisy scene tokens and fed into the transformer via a shared self-attention stream. We optimize the flow-matching objective in Eq.~\ref{eq:flow_matching_compact} using AdamW, with a learning rate of $10^{-4}$, weight decay of $0.01$, 5k warm-up steps, and cosine decay toward $10^{-5}$. We apply conditional dropout with a probability of 0.1, maintain an exponential moving average (EMA) with a decay of 0.9999, and clip gradients to unit norm. The reported generator is trained for 100k steps. 

\paragraph{Generation Inference and Evaluation.}
\label{supp:generation_metric}
For all generation experiments, we use the pretrained parameters of the
latent flow model. At inference, we integrate the learned flow from
$\sigma=1$ to $\sigma=0$ using UniPC with 50 function evaluations
and a classifier-free guidance scale of $1.0$. For
single-image-conditioned generation, each held-out test scene is conditioned on one input image, which provides a partial observation of the complete 3D scene. We generate one sample per condition using fixed random seeds. Generated and reference 3DGS scenes are rendered
from the same 16 evaluation cameras at a resolution of
$720\times480$.

\textbf{Appearance metrics.}
We evaluate rendered appearance using Fr\'echet Inception Distance (FID) and Kernel Inception Distance (KID), computed from standard Inception-v3 features using the PyTorch-based metric implementation.
For single-image conditioning, we separately evaluate regions observed and unobserved from the conditioning view. Specifically, we back-project valid pixels of the conditioning image into 3D using the corresponding
depth and camera parameters, and reproject these 3D samples into each evaluation view. The resulting projection footprint defines the observed region, while the remaining valid region is treated as
unobserved. The same region masks are applied to the generated and reference renderings when computing FID and KID. The masks are used exclusively for evaluation and are never provided to the generative model.

\textbf{Geometry metrics.}
We evaluate geometric generation using Coverage (COV) and Minimum Matching Distance (MMD) under Chamfer Distance (CD). Each generated and reference scene is represented by 2,048 points sampled from its Gaussian centers. For single-image conditioning, we use the same
geometric projection procedure to distinguish observed and unobserved scene geometry. Specifically, the conditioning observation is lifted into 3D using its depth and camera parameters, and its projection,
together with the corresponding projected depth, determines the portion of the scene covered by the input observation. Geometry outside this observed support is treated as unobserved. We sample 2,048 points
from the corresponding unobserved geometry and compute COV-CD and MMD-CD between generated and reference scene sets under the same spatial normalization.

\subsection{Additional VAE Reconstruction Results}
\label{sec:supp_vae_qualitative}

Figure~\ref{fig:vae_comp_supp} provides additional qualitative comparisons with GaussianGPT. In all three scenes, our method can preserve both the global room layout and fine-grained local structures. While GaussianGPT captures the coarse spatial envelope, it struggles with thin structures and details, often producing blurred or fragmented geometry. Although our reconstructions are not flawless, regions with sparse observations near scene boundaries appear less sharp than the ground truth; they still preserve the high-fidelity scene-specific geometry and appearances.

\begin{figure}[t]
    \centering
    \includegraphics[width=\linewidth]{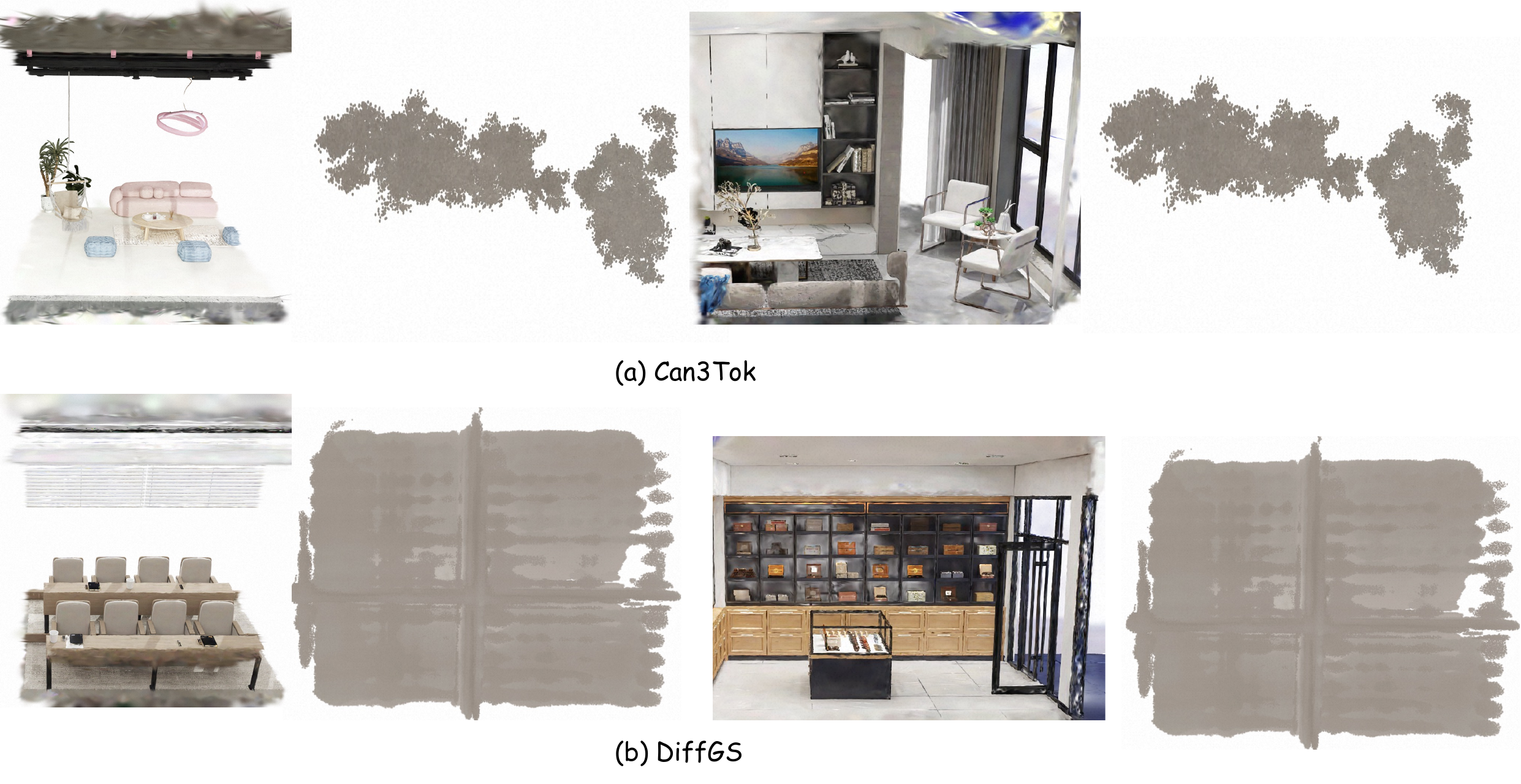}
    \caption{\textbf{Representative reconstruction failures of the adapted Can3Tok and DiffGS
baselines.}  Both baselines are trained using their officially released implementations on the same scene-level training split and with the same optimization budget used in our comparison. Under this setting, distinct input scenes can be reconstructed as similar low-detail structures. These examples illustrate the observed behavior under our scene-level adaptation and are not intended to characterize the methods outside this
setting.}
    \label{fig:failure_case}
\end{figure}


Figure~\ref{fig:failure_case} further examines a representative failure mode of the adapted Can3Tok and DiffGS baselines. Both methods tend to decode visually distinct input scenes into similar, low-detail Gaussian structures. They recover only a rudimentary spatial envelope, discarding most of the scene-specific layouts, appearances, and object structures.  We present these cases as diagnostic examples to understand representational limits, rather than as a universal characterization of either baseline.

\begin{table}[t]
  \centering
  \setlength{\tabcolsep}{3pt}
  \begin{minipage}[t]{0.48\columnwidth}
    \centering
    \caption{\textbf{Local Gaussian matching.}
    We compare how predicted and target Gaussians are matched inside each decoded voxel.}
    \vspace{-0.1cm}
    \label{tab:abl_assignment}
    \footnotesize
    \begin{adjustbox}{max width=\linewidth}
    \begin{tabular}{lcccc}
      \toprule
      Setting & PSNR$\uparrow$ & SSIM$\uparrow$& LPIPS$\downarrow$  & CD$\downarrow$ \\
      \midrule
      Hard matching  & 25.61   & \textbf{0.908}  & 0.103 & 2.29 \\
      Sinkhorn OT (Ours) & \textbf{25.83} & {0.904} & \textbf{0.096} & \textbf{2.27} \\
      \bottomrule
    \end{tabular}
    \end{adjustbox}
  \end{minipage}
  \hfill
  \begin{minipage}[t]{0.48\columnwidth}
    \centering
    \caption{\textbf{Triplane resolution.}
    $R_p$ denotes the plane-grid resolution prior to latent compression.}
    \vspace{-0.1cm}
    \label{tab:abl_resolution}
    \footnotesize
    \begin{adjustbox}{max width=\linewidth}
    \begin{tabular}{lcccc}
      \toprule
      Setting & PSNR$\uparrow$ & SSIM$\uparrow$ & LPIPS$\downarrow$ & CD$\downarrow$ \\
      \midrule
      $R_p=64$  & 21.07 & 0.860 & 0.155 & 2.65 \\
      $R_p=128$ & 24.15 & 0.882 & 0.118 & 2.42 \\
      $R_p=256$ (Ours) & \textbf{25.83} & \textbf{0.904} & \textbf{0.096} & \textbf{2.27} \\
      \bottomrule
    \end{tabular}
    \end{adjustbox}
  \end{minipage}
\end{table}

\subsection{Expanded Ablation Results and Analysis}
\label{sec:supp_ablation}

\begin{figure}[t]
    \centering
    \includegraphics[width=\linewidth]{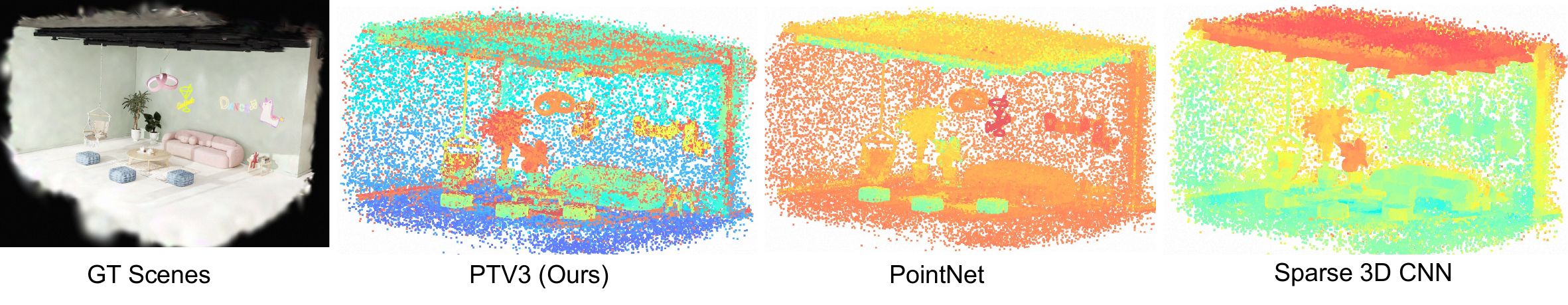}
    \caption{\textbf{Encoder feature comparison.} We visualize features extracted by PTv3, PointNet, and a sparse 3D CNN from the same Gaussian input. PTv3 retains clearer spatial variation around surfaces, objects, and scene boundaries in this example; Table~\ref{tab:abl_writer_reader} reports the corresponding quantitative comparison.}
    \label{fig:encoder_comparison}
\end{figure}

\begin{figure}[t]
    \centering
    \includegraphics[width=\linewidth]{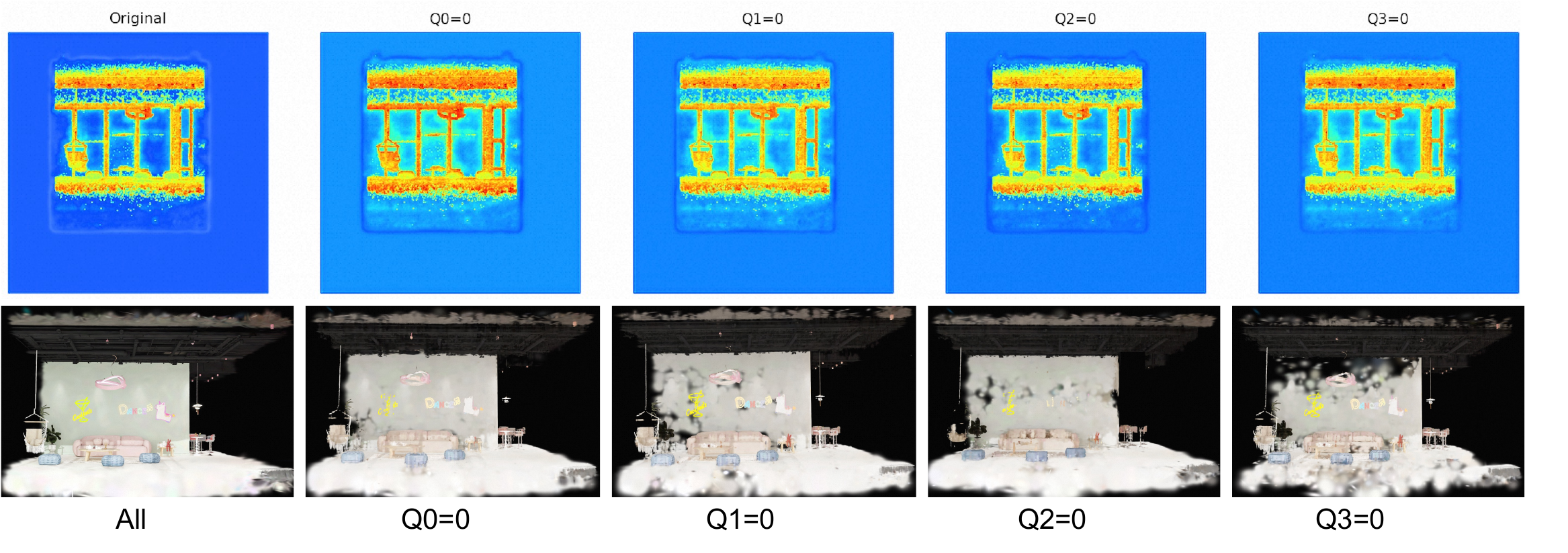}
    \caption{\textbf{Single-slot removal.} This intervention complements the slot-count ablation in Table~\ref{tab:abl_slots}. We zero one of the $Q$ encoder slots while keeping all remaining latent values fixed. The top row visualizes decoded feature energy on the $xz$ plane, and the bottom row shows the corresponding Gaussian reconstruction. Different removals produce distinct changes, indicating that the slots are not identical copies of the same local summary.}
    \label{fig:q_token_ablation}
\end{figure}

We complement the component ablations in the main paper with controlled studies on local Gaussian matching and representation capacity. Tables~\ref{tab:abl_assignment} and~\ref{tab:abl_resolution} examine the matching objective and triplane resolution, while Tables~\ref{tab:abl_slots} and~\ref{tab:abl_latent_width} vary the number of encoder slots and latent channels. Figures~\ref{fig:encoder_comparison} and~\ref{fig:q_token_ablation} provide qualitative evidence supporting the necessity of contextual point features and multiple local slots.

For all reconstruction ablations, PSNR, SSIM, and LPIPS are computed on the same held-out renderings, while Chamfer Distance (CD) is evaluated on decoded Gaussian geometry under identical normalization. Each hyperparameter variant is trained from scratch using the same data and evaluation split. 

Figure~\ref{fig:encoder_comparison} highlights the importance of the point backbone prior to triplane projection. PointNet~\cite{qi2017pointnet} extracts largely uniform features across the scene, resulting in limited local discriminability. While the sparse 3D CNN captures broad spatial variations, these variations mostly correspond to coarse, scene-level regions. In contrast, PTv3 features exhibit sharp transitions around floors, walls, furniture, and object boundaries. This preserves critical local structures, allowing the encoder to more accurately assign features to triplane cells. This qualitative observation aligns with the architectural comparisons in Table~\ref{tab:abl_writer_reader} and justifies contextualizing Gaussian primitives before local aggregation.

\begin{table}[t]
  \centering
  \setlength{\tabcolsep}{4pt}
  \begin{minipage}[t]{0.48\columnwidth}
    \centering
    \caption{\textbf{Encoder token slots.}
    $Q$ denotes the number of feature slots per triplane cell.}
    \label{tab:abl_slots}
    \begin{adjustbox}{max width=\linewidth}
    \begin{tabular}{l|cccc}
      \toprule
      Setting & PSNR$\uparrow$ & SSIM$\uparrow$ & LPIPS$\downarrow$ & CD$\downarrow$ \\
      \midrule
      $Q=1$         & 24.40 & 0.883 & 0.114 & 2.38 \\
      $Q=2$         & 25.26 & 0.895 & 0.105 & 2.32 \\
      $Q=4$ (Ours) & 25.83 & 0.904 & 0.096 & 2.27  \\
      $Q=8$         & \textbf{26.11} & \textbf{0.910} & \textbf{0.090} & \textbf{2.23} \\
      \bottomrule
    \end{tabular}
    \end{adjustbox}
  \end{minipage}
  \hfill
  \begin{minipage}[t]{0.48\columnwidth}
    \centering
    \caption{\textbf{Latent channel width.}
    $C_z$ specifies the number of latent channels per plane.}
    \label{tab:abl_latent_width}
    \begin{adjustbox}{max width=\linewidth}
    \begin{tabular}{lcccc}
      \toprule
      Setting & PSNR$\uparrow$ & SSIM$\uparrow$ & LPIPS$\downarrow$ & CD$\downarrow$ \\
      \midrule
      $C_z=8$         & 25.16 & 0.892 & 0.102 & 2.34 \\
      $C_z=16$ (Ours) & 25.83 & 0.904 & 0.096 & 2.27 \\
      $C_z=32$        & \textbf{25.89} & \textbf{0.905} & \textbf{0.094} & \textbf{2.25} \\
      \bottomrule
    \end{tabular}
    \end{adjustbox}
  \end{minipage}
\end{table}

\paragraph{Local Matching and Spatial Resolution.}
Table~\ref{tab:abl_assignment} reveals a metric-dependent trade-off between hard assignment and Sinkhorn OT. Sinkhorn improves PSNR from 25.61 to 25.83, reduces LPIPS from 0.103 to 0.096, and reduces CD from 2.29 to 2.27, while hard matching yields a slightly higher SSIM (0.908 versus 0.904). Overall, the transport-based objective yields superior perceptual and geometric reconstructions, even if it does not dominate every single metric. Table~\ref{tab:abl_resolution} exhibits a clearer trend: increasing $R_p$ from 64 to 256 improves PSNR by 4.76 dB and SSIM by 0.053, while reducing LPIPS by 0.059 and CD by 0.38. These consistent gains demonstrate that a coarse grid prematurely discards spatial details prior to latent compression.

\paragraph{Slot Capacity and Latent Width.}
Increasing the encoder capacity from one to four slots improves all four metrics (Table~\ref{tab:abl_slots}), including a 1.43 dB increase in PSNR and a 0.11 reduction in CD. This confirms that utilizing multiple tokens per projected cell is crucial when multiple Gaussian primitives collapse into the same local region. Increasing the slot count from four to eight yields further but diminishing gains, revealing a quality--capacity trade-off. By comparison, widening the latent representation from $C_z=16$ to $C_z=32$ changes PSNR by only 0.06 dB, SSIM by 0.001, LPIPS by 0.002, and CD by 0.02 (Table~\ref{tab:abl_latent_width}). This indicates that the representation benefits significantly more from structured local capacity (slots) than from a brute-force increase in latent channels. We use $Q=4$ and $C_z=16$ as a compact operating point; however, $Q=8$ remains a viable option when reconstruction quality is strictly prioritized over representation size.

\begin{wraptable}{r}{0.50\linewidth}
  \centering
  \vspace{-0.4cm}
  \caption{
  Effect of local Gaussian decoder capacity.
  $K$ is the maximum number of Gaussian candidates per selected voxel,
  while rendering supervision always uses the complete target scene.
  CD values are multiplied by $10^3$.
  Best results are in \textbf{bold}, second-best are \underline{underlined}.
  }
  \vspace{-0.25cm}
  \label{tab:decoder_capacity}
  \footnotesize
  \setlength{\tabcolsep}{4pt}
  \begin{tabular}{lcccc}
    \toprule
    Setting & PSNR$\uparrow$ & SSIM$\uparrow$
            & LPIPS$\downarrow$ & CD$\downarrow$ \\
    \midrule
    $K=8$
        & 25.06 & 0.887 & 0.112 & 2.31 \\
    $K=16$
        & 25.61 & 0.895 & 0.106 & \underline{2.33} \\
    \rowcolor{cyan!10}
    $K=24$ (Ours)
        & \underline{25.83} & \underline{0.904}
        & \underline{0.096} & \textbf{2.27} \\
    $K=32$
        & \textbf{25.96} & \textbf{0.907}
        & \textbf{0.092} & \textbf{2.27} \\
    \bottomrule
  \end{tabular}
  \vspace{-8pt}
\end{wraptable}
\paragraph{Local Gaussian Decoder Capacity.}
We further study the effect of the maximum number of Gaussian candidates
$K$ predicted within each selected voxel. $K$ defines a local capacity
upper bound rather than a fixed scene-level output cardinality: each
selected voxel retains a content-dependent subset of the $K$ candidates,
while the number and spatial distribution of selected voxels remain
scene-dependent. As shown in Table~\ref{tab:decoder_capacity}, increasing
$K$ from 8 to 24 generally improves reconstruction quality, with
particularly consistent gains in the appearance metrics. Increasing the
capacity further to $K=32$ yields only marginal improvements
($+0.13$ dB PSNR, $+0.003$ SSIM, and $-0.004$ LPIPS), while CD remains
unchanged at the reported precision. These results indicate that
$K=24$ captures most of the useful local representation capacity in our
setting.

This behavior is consistent with the local-density statistics in
Figure~\ref{fig:stats_3dgs}: $97.8\%$ of occupied voxels in the original
fitted scenes contain at most 24 Gaussians on the same $256^3$ grid.
For the remaining high-density tail, the set-level correspondence target
is truncated to at most $K$ primitives, whereas the rendering target
remains untruncated and is computed from the complete fitted Gaussian
scene before local primitive selection. Consequently, although the
decoder cannot reproduce more than $K$ individual primitives within a
single voxel, the retained Gaussians can jointly adjust their non-positional attributes under full-scene rendering supervision to
approximate the image-space effect of the denser target representation.
The small performance gap between $K=24$ and $K=32$ empirically suggests
that this bounded local capacity is not a major bottleneck for
reconstruction fidelity under our data distribution.

\begin{figure}[t]
    \centering
    \vspace{-0.2cm}
\includegraphics[width=\linewidth]{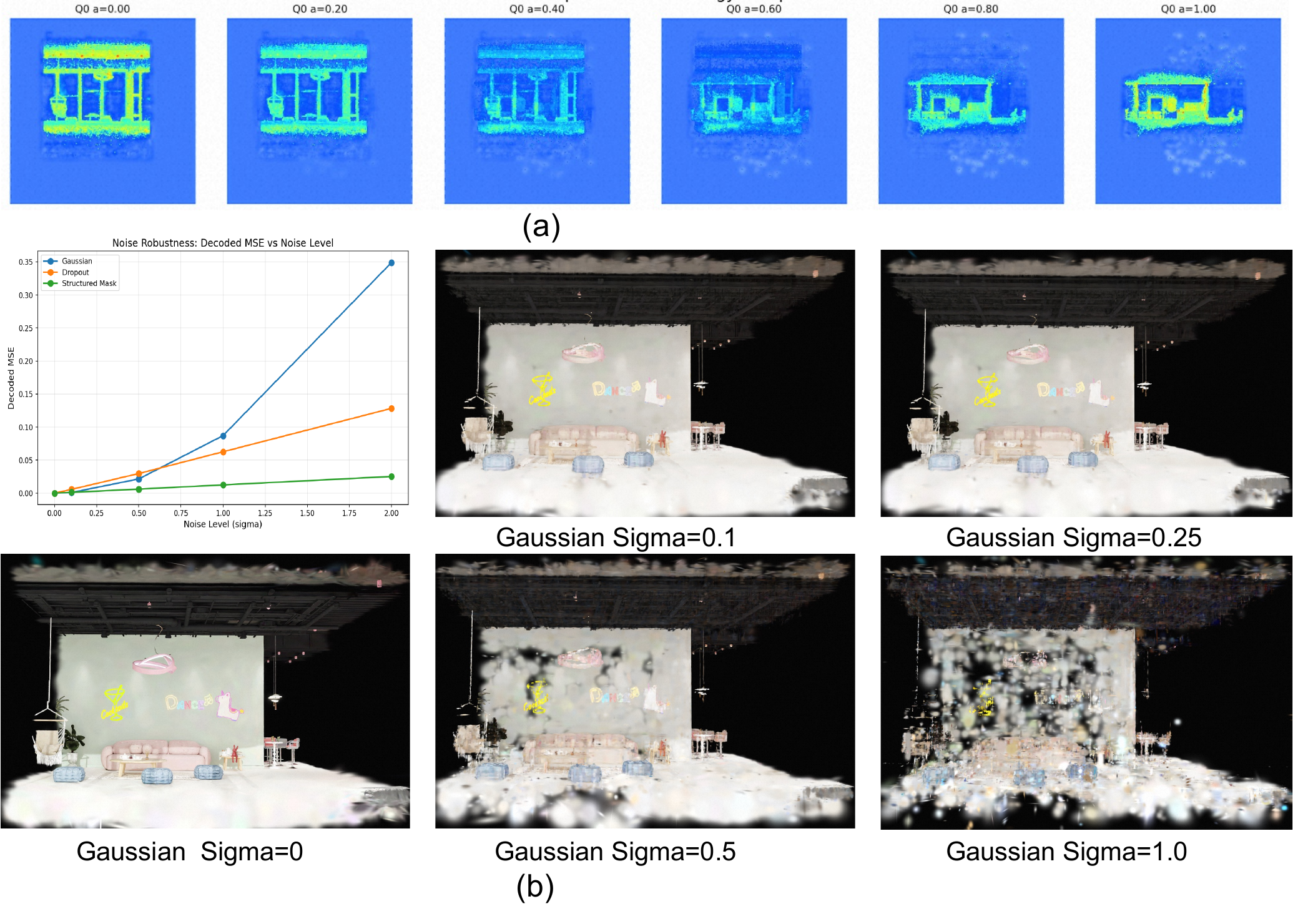}
    \vspace{-0.2cm}
    \caption{\textbf{Latent interpolation and perturbation analysis.}
    (a) Linear interpolation produces smooth changes in decoded triplane features.
    (b) The latent representation remains stable under moderate perturbations, with
    pronounced degradation under Gaussian noise at $\sigma=1.0$.}
    \label{fig:latent_robustness}
    \vspace{-0.8cm}
\end{figure}

Figure~\ref{fig:q_token_ablation} probes the four-slot representation without retraining the decoder. While masking out any single slot keeps the global room layout recognizable, it alters the feature response and degrades distinct local structures in the reconstructed Gaussians. The degraded regions and artifact patterns vary depending on which slot is removed, rather than presenting a uniform failure mode. Together with the monotonic slot-count improvements in Table~\ref{tab:abl_slots}, this demonstrates that the slots learn complementary local information. While this does not imply a strictly disentangled semantic meaning for each slot, it strongly supports the presence of non-redundant local decomposition within the projected cells.


\subsection{Latent Representation Analysis}
\label{sec:latent_analysis}

To evaluate the structural integrity and generative suitability of the learned VAE latent space, we analyze its smoothness and robustness through interpolation and perturbation. For interpolation, we linearly interpolate between two distinct scene latents, $\mathbf{z}_{\alpha}=(1-\alpha)\mathbf{z}_A+\alpha\mathbf{z}_B$. As shown in Fig.~\ref{fig:latent_robustness}(a), the decoded triplane features exhibit smooth, continuous semantic transitions along the interpolation path, confirming a well-regularized latent manifold.

To assess robustness, we systematically inject Gaussian noise, apply element-wise dropout, and enforce structured masking on the latent codes, subsequently quantifying the resulting degradation relative to the unperturbed reconstruction. We further visualize the decoded triplane features under increasing masking ratios, where selected latent components are zeroed out. As shown in Fig.~\ref{fig:latent_robustness}(b), the representation remains highly resilient under moderate perturbations: the decoded scene is largely preserved under Gaussian noise up to $\sigma=0.5$, gracefully degrading only at extreme noise levels ($\sigma=1.0$).

\subsection{Additional Conditional Generation Results}
\label{sec:supp_more_generation}

We provide additional qualitative results for both point-cloud- and single-image-conditioned generation. These visual examples complement the quantitative evaluations in Sec.~\ref{sec:exp_generation} by demonstrating the structural diversity of the generated 3D Gaussian scenes.

\begin{figure}[!ht]
    \centering
    \includegraphics[width=\linewidth]{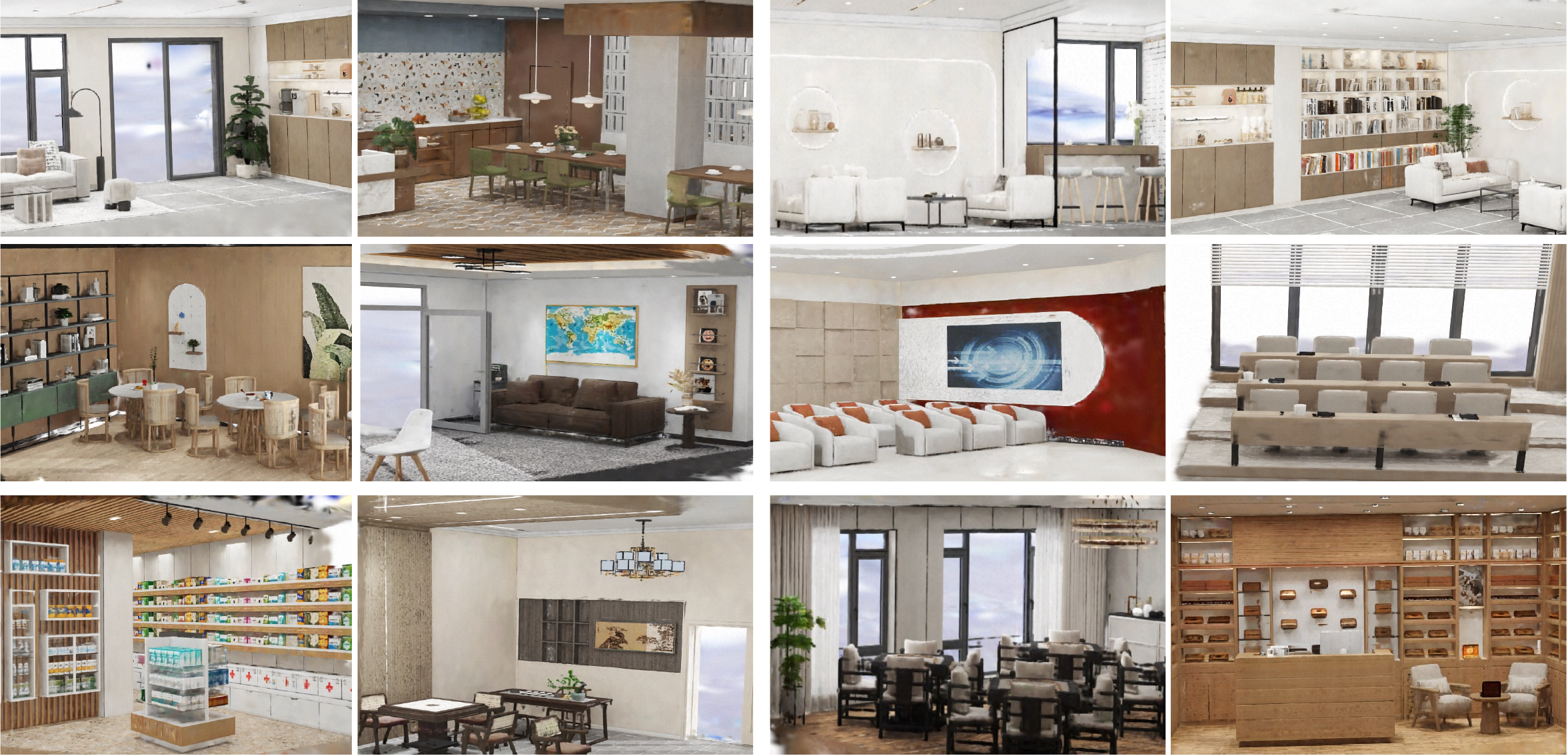}
    \caption{\textbf{Additional point-cloud-conditioned generations.} Each panel shows a Gaussian scene sampled from a different point-cloud condition. The examples illustrate the diversity of generated indoor layouts and scene content.}
    \label{fig:pcd2gs_supp}
\end{figure}

\paragraph{Point-cloud Conditioned Generation.}
Figure~\ref{fig:pcd2gs_supp} presents supplementary scenes generated from full point-cloud conditions. These samples span a wide variety of indoor layouts that encompass diverse arrangements of structures. Notably, the generated Gaussian sets successfully assemble into holistic, scene-scale environments rather than degenerating into isolated objects or disconnected fragments. The results show that major surfaces and semantic arrangements are robustly generated. 

\begin{figure}[!ht]
    \centering
    \includegraphics[width=\linewidth]{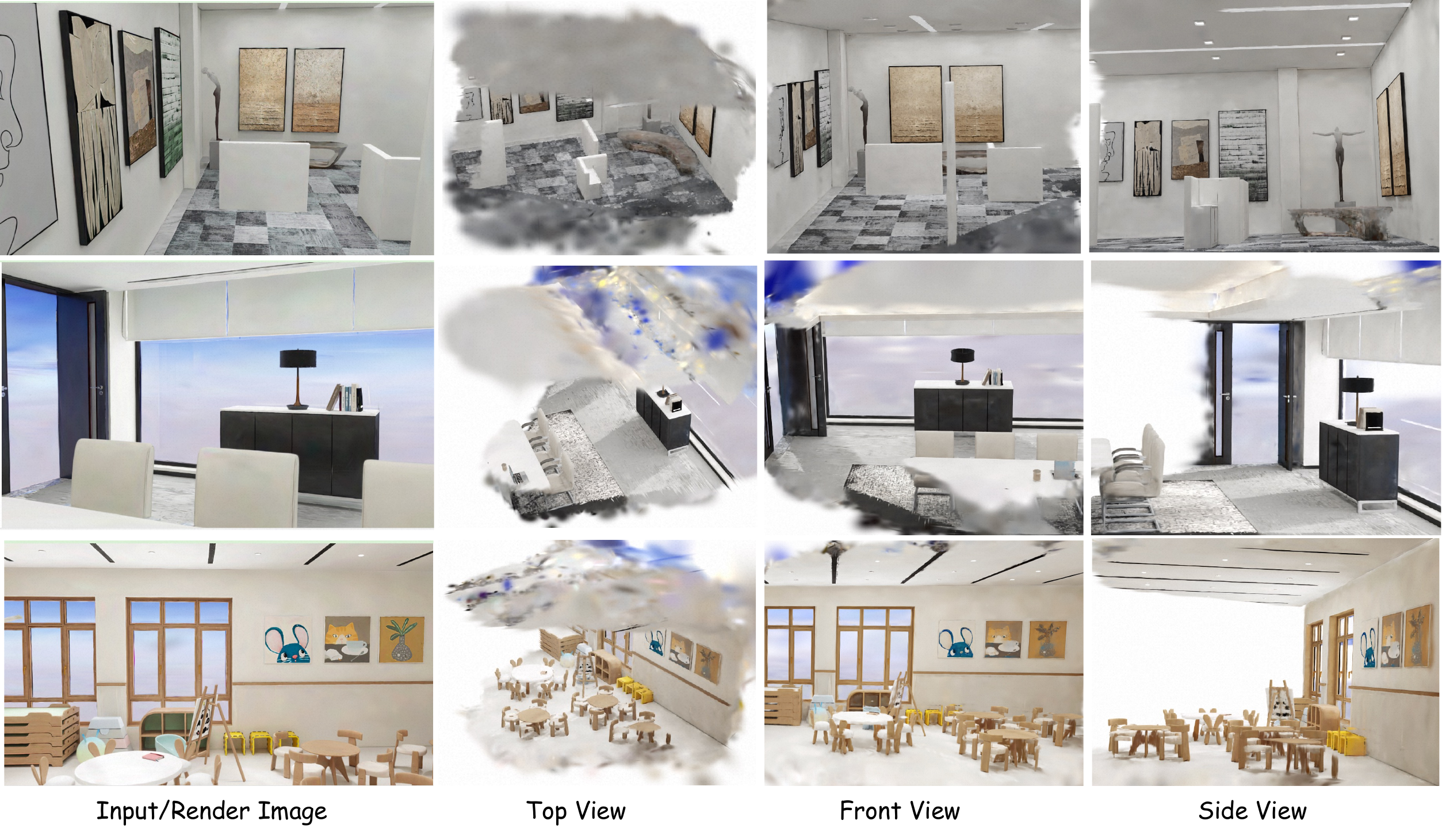}
    \caption{\textbf{Single-image-conditioned generation.} Each row shows the conditioning view followed by top, front, and side renderings of one sampled Gaussian scene. The novel views assess whether input-specific scene cues are preserved within a coherent 3D completion.}
    \label{fig:image2gs_supp}
\end{figure}

\paragraph{Single-image Conditioned Generation.}
Figure~\ref{fig:image2gs_supp} shows some examples of single-image-conditioned generation. In each row, the input view is followed by top, front, and side renderings of a single sampled Gaussian scene. The generated scenes preserve prominent cues from the input, including wall decorations, window placements, cabinets, and furniture configurations. Crucially, these elements remain geometrically consistent across novel viewpoints: planar surfaces maintain their relative spatial arrangements, and scattered objects (e.g., tables and chairs) are grounded coherently within the shared 3D space.

\paragraph{Limitations.}
\label{supp:limitation}
Our current framework is primarily limited by the scale and diversity of
available scene-level 3DGS training data. In contrast to image and video
VAEs and generative models, which can benefit from substantially larger and
more diverse training corpora, our Gaussian VAE and latent flow model are
trained on a relatively limited collection of 3D scenes. This restricts the
diversity of the learned scene prior and may limit generalization to scene
distributions that differ substantially from the training data. We expect
larger and more diverse 3DGS datasets to further improve both representation
learning and generative modeling. A second limitation concerns physical
scene scale. To obtain a consistent spatial coordinate system, we construct
all training samples using the same fixed-size 3D crop and map them to a
canonical triplane domain. While this fixed-scale normalization stabilizes
representation learning, it also ties the learned representation to the
physical scale used during training, and scenes at substantially different
spatial scales may not be represented optimally without additional
normalization. Extending \modelname{} to scale-conditioned or multi-scale
representations, together with training on larger 3DGS corpora, is therefore
an important direction for improving its generalization to more diverse
real-world scenes.

\end{document}